\pdfoutput=1
\documentclass[11pt]{article}

\usepackage[preprint]{acl}

\usepackage{times}
\usepackage{latexsym}
\usepackage[T1]{fontenc}
\usepackage[utf8]{inputenc}
\usepackage{microtype}
\usepackage{inconsolata}

\usepackage{amsmath,amssymb}
\usepackage{booktabs}
\usepackage{array}
\usepackage{multirow}
\usepackage{graphicx}
\usepackage{xcolor}
\usepackage{colortbl}
\DeclareRobustCommand{\framework}{\mbox{\textcolor{MethodAccent}{Cache2LoRA}}}
\DeclareRobustCommand{\ours}{\textcolor{MethodAccent}{Cache2LoRA}}
\DeclareRobustCommand{\ourstf}{\mbox{\textcolor{MethodAccent}{Cache2LoRA-TF}}}
\definecolor{MethodAccent}{HTML}{245E63}
\definecolor{PaperHeader}{HTML}{C9D7E8}
\definecolor{PaperGroup}{HTML}{EDF5F8}
\definecolor{PaperBest}{HTML}{DCEDEA}
\DeclareRobustCommand{\methodname}{\textcolor{MethodAccent}{\textbf{\ours}}}
\newcommand{\tablelead}[1]{\textbf{#1}}
\DeclareRobustCommand{\methodnametf}{\textbf{\ourstf}}

\title{\ours{}: Understanding and Improving Context Internalization}

\author{
  Yi Fang$^{1}$ \quad
  Gaurav Srivastava$^{1}$ \quad
  Zhenyu Bi$^{1}$ \quad
  Sirui Ding$^{2}$ \quad
  Xuan Wang$^{1}$\thanks{Corresponding author: \texttt{xuanw@vt.edu}.} \\
  $^{1}$Department of Computer Science, Virginia Tech, Blacksburg, VA, USA \\
  $^{2}$Department of Biomedical Data Science, Stanford University, Stanford, CA, USA \\
  \texttt{\{yif,gks,zhenyub,xuanw\}@vt.edu} \\
  \texttt{srding@stanford.edu}
}

\begin{document}
\maketitle

\begin{abstract}
Context-to-LoRA methods encode documents as low-rank parameter updates, allowing language models to answer without retaining the source text in their decoding context. How reliably do these updates preserve facts and support their use under different instructions? Controlled diagnosis of existing context-to-LoRA methods separates fact discrimination, target retention, and response adaptation. Competing sources influence answers, while target-only construction and memory-state optimization improve recovery at fixed update size. Recovering facts also does not ensure that answers follow response instructions. We introduce \framework{}, which uses each query to aggregate evidence from a reusable frozen document cache before generating LoRA updates. Trainable encoding operates only on the query and memory-token suffix, and multi-style supervision teaches both evidence preservation and answer expression. Across 17 tasks, \ours{} improves on existing context-to-LoRA methods and achieves macro F1 of 0.663, approaching full-context in-context learning at 0.671. For 32K documents and 256-token outputs at batch size 16, its decoding throughput is 5.9$\times$ that of full-context ICL. At matched memory size, generated LoRA preserves answer accuracy more reliably than the learned-KV variant under long, diverse intervening text and combines more effectively with retained document KV, yielding higher hybrid \mbox{F1 and EM}.
\end{abstract}

\section{Introduction}
Parameter memory represents document information as weight updates rather than additional tokens in the answering context. Recent writers generate LoRA updates in a forward pass, avoiding per-document gradient optimization \citep{charakorn2026d2l,liu2026shine}. This offers a way to carry document information into the model's computation without attending to document positions during decoding. Can these generated updates preserve the right facts and support their use under different response instructions?

We diagnose D2L and SHINE along three dimensions: fact discrimination, target retention, and response adaptation. Our source-substitution experiments link borrowed answers to competing facts. When we restrict writing to the passage relevant to the question, factual recovery improves without increasing the update size. Optimizing the writer's memory states also improves recovery at fixed output size. These controls suggest that the difficulty lies partly in determining which information to write, beyond how much the update can hold. We also find that D2L can recover reference content without adopting the requested answer form. Effective internalization requires selecting relevant facts and learning to express them as instructed.

For fact selection, the query supplies information that document-only writers lack: what the reader needs from the document. This motivates using the query during writing, but the resulting construction cost must be paid for every question. Within parameter memory, Doc-to-Atom enables query routing but requires semantic decomposition and compilation into LoRA units beforehand \citep{diao2026atom}. AAG first generates a synthetic document and uses it with the query to produce adapters, coupling parameter construction to an additional context-generation stage \citep{liao2025awakening}. Activation-based compressors instead produce compact embeddings or KV states. OSCAR, QGC, and ATACompressor jointly encode document content with each query, repeating document encoding when another question concerns the same source \citep{louis2025oscar,cao2024qgc,li2025atacompressor}.

We propose \textbf{\framework{}}, which generates query-specific LoRA directly from a reusable cache of the supplied document. A frozen backbone prefills the document once, without semantic atomization or synthetic-context generation. For each question, trainable encoding acts only on the query and appended memory tokens, aggregating relevant information without re-encoding the document. A generation head converts these states into LoRA updates for the answerer. Multi-style teacher supervision supports both factual recovery and varied response instructions, while a retention objective limits interference on unrelated inputs. New documents and queries require no gradient optimization.

Across 17 tasks, \ours{} reaches macro F1 of 0.663, approaching full-context ICL's 0.671, and improves on released D2L, \ourstf{}, and further-trained D2L. Cached construction avoids repeated document encoding. For 32K documents and 256-token outputs at batch size 16, decoding throughput reaches 5.9$\times$ that of full-context ICL. Generated LoRA also shows two advantages over directly using the learned memory tokens as KV states at matched memory size. First, it preserves answer accuracy more reliably under long, diverse intervening text, keeping query-specific document memory usable after construction. Second, it complements retained document KV more effectively, yielding higher hybrid F1 and EM across both selectors and all tested budgets.

\begin{figure*}[!t]
\centering
\includegraphics[width=\textwidth]{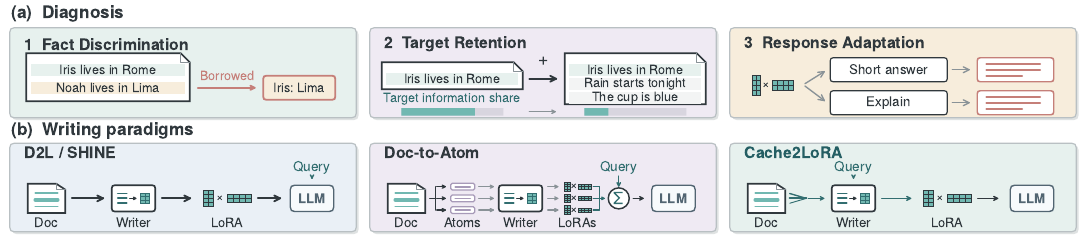}
\caption{Illustrative diagnostics and writing paradigms. (a) Fact discrimination, target retention, and response adaptation assess factual recovery and answer expression. Teal bars schematize relative target shares. (b) Queries guide answering in all three paradigms, memory routing in Doc-to-Atom, and memory writing in Cache2LoRA.}
\label{fig:overview}
\end{figure*}

\section{Related Work}

\subsection{Context-to-LoRA Generation}
Context distillation transfers source-conditioned behavior into parameters \citep{snell2022context,caccia2025deep}, while hypernetworks predict updates from task or input representations \citep{ha2016hypernetworks,mahabadi2021hyperformer,ivison2022hyperdecoders,phang2022hypertuning}. Our writer combines generated adaptation with distillation. D2L maps document latents to LoRA, and SHINE maps memory-token states to updates \citep{charakorn2026d2l,liu2026shine}. Both write before seeing the query. Our diagnoses of factual recovery and response adaptation motivate query-guided writing.

Parametric RAG and Doc-to-Atom select precompiled parameter memories \citep{su2025prag,diao2026atom}, whereas we generate an update for each query. AAG constructs adapters from synthetic context \citep{liao2025awakening}, while we use the supplied document's cache. Override adjusts generated updates to resolve conflicts with pretrained knowledge \citep{cheng2026override}. We instead train the writer to select facts and express answers.

\subsection{Fast Weights and Test-Time Adaptation}
Fast-weight programmers write input-dependent weights, encompassing certain linear-attention updates \citep{schlag2021linear}. Our writer likewise generates dynamic parameters, specifically LoRA updates for a pretrained answerer. Test-time training optimizes parameters during inference \citep{sun2020ttt,hardt2024tttnn}. TTT layers use a trainable model as recurrent state, and Titans updates neural memory at test time \citep{sun2024tttlayers,behrouz2025titans}. \ours{} instead generates updates from cached document states without test-document gradients. Forward generation is also used in fast-weight systems.

\subsection{Context Compression}
Text selection and learned activation memories provide alternatives to parameter storage \citep{jiang2024longllmlingua,mu2023gist,ge2023icae,chevalier2023autocompressors}. We use CCM and 500xCompressor as KV controls \citep{kim2024ccm,li2025compressor}. OSCAR, QGC, and ATACompressor condition compression on the query \citep{louis2025oscar,cao2024qgc,li2025atacompressor}, but jointly encode document content with each question. Our writer reuses document states. We also compare with released OSCAR on Mistral.

KV eviction selects existing states \citep{zhang2023h2o,xiao2024streaming,li2024snapkv,zhang2026dropkv,kim2026fastkvzip}, whereas our writer aggregates document information into new parameter updates. Cartridges and related studies compare KV and parameter memories optimized for a corpus or document \citep{eyuboglu2026cartridges,rakotonirina2026document}. Our comparison instead uses forward-generated, query-specific memories.

\section{Diagnosis}
\label{sec:diagnosis}
We examine fact discrimination, target retention, and response adaptation in D2L, and repeat selected tests on SHINE (Figure~\ref{fig:overview}). The base model answers with the question alone or with the full document through in-context learning (ICL). Appendices~\ref{app:baselines} and~\ref{app:diagnostic-protocols} detail the baselines and each test's data, procedures, controls, and metrics.

\subsection{Fact Discrimination}
\textbf{D2L can confuse facts from different sources.} We combine Wikipedia passages about different people and ask for one person's birth year. Each passage names the person and birth year, so we can trace a wrong answer to another person. Tests that vary the number of people also cover D2L on Gemma and SHINE on Qwen (Figure~\ref{fig:aliasing-fact-count}).

\textbf{Changing the borrowed fact changes the answer.} In 67 responses, the first reported year belongs to another person. We change that person's year in the source text, regenerate the LoRA update, and ask the same question. The first year changes in 91\% of cases, versus 27\% when we change a third person's year. Averaging the vectors read from the borrowed passage during writing changes the first year in 65/67 cases, versus 22/67 for a third passage. Borrowed and other passages receive similar mean attention shares (0.119 vs.\ 0.117), so attention share alone does not identify which fact is borrowed (Appendix~\ref{app:aliasing-internals}).

\begin{figure}[!tb]\centering
\includegraphics[width=\linewidth]{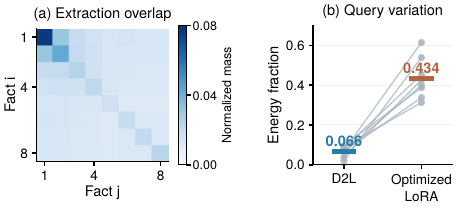}
\caption{Internal readouts on FactSet-Homo. (a) Overlap between facts' extraction contributions over 16 documents. (b) Variation in LoRA's direct effect across questions, measured as a fraction of activation energy. Lines pair D2L and directly optimized LoRA of the same size on eight documents. Thick marks show means.}
\label{fig:mode1}
\end{figure}

\textbf{D2L's direct contribution varies little across questions.} We hold each document's adapter fixed and measure its contribution to activations at a fixed answer prefix across eight questions. The fraction of activation energy that varies across questions averages 0.066 for D2L versus 0.434 for directly optimized LoRA of the same size (Figure~\ref{fig:mode1}b). Figure~\ref{fig:mode1}a also shows overlap between facts' contributions during writing. This overlap is only weakly correlated with changes in competing-answer scores ($r=0.057$, Appendix~\ref{app:aliasing-internals}).

\subsection{Target Retention}
\textbf{Adding other content makes the requested fact harder to recover.} In FactSet, we keep the question and the fact needed to answer it fixed while adding statements. ROUGE-L Recall (R-L) measures how much of the reference answer appears in the response. On FactSet-Homo, D2L scores 0.982 when LoRA is generated from that fact alone and 0.184 when other facts are included, while ICL retains full coverage (Table~\ref{tab:mode1}). Generic sentences without names or numbers cause smaller losses than added facts (Appendix~\ref{app:dilution-internals}). SQuAD and NQ mark the passages containing each answer, allowing the same comparison on natural text (Table~\ref{tab:mode1}).

\begin{figure}[!tb]\centering
\includegraphics[width=\linewidth]{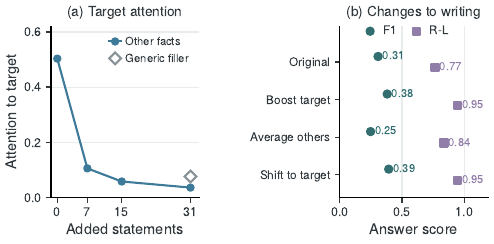}
\caption{Changes to writing on 64 questions. (a) Attention share given to the target as statements are added. (b) F1/R-L with 15 added facts: unchanged writing, boosting target attention, averaging other statements' vectors, or shifting attention toward the known target.}
\label{fig:mode2}
\end{figure}

\begin{figure}[!tb]\centering
\includegraphics[width=\linewidth]{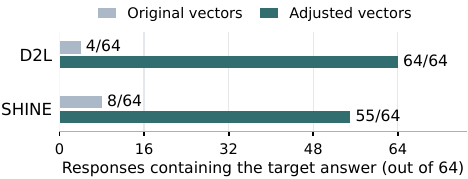}
\caption{Adjusting the vectors used to generate LoRA. The answerer, generator, and LoRA size stay fixed. Supervised training changes only the generator's input vectors. Bars count responses containing an accepted answer out of 64 questions (Appendix~\ref{app:frozen-head}).}
\label{fig:capacity-control}
\end{figure}

\textbf{Changing what is written improves recovery at the same LoRA size.} In a 64-question test, we track the attention given to the target statement as more facts are added (Figure~\ref{fig:mode2}a). Increasing only the target's attention weights, with all added facts still present, raises R-L from 0.77 to 0.95 with 15 additions (Figure~\ref{fig:mode2}b). We also freeze the answerer and LoRA generator, then use supervised training to adjust only the vectors passed to the generator. At the same LoRA size, responses containing the correct answer increase from 4/64 to 64/64 for D2L and from 8/64 to 55/64 for SHINE (Figure~\ref{fig:capacity-control}). \textbf{On these examples, the fixed generator can produce more effective LoRA updates, pointing to the construction of its input vectors as a bottleneck.}

\subsection{Response Adaptation}
\begin{figure}[!tb]\centering
\includegraphics[width=\linewidth]{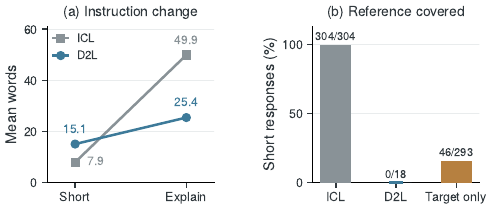}
\caption{Answer form under changed instructions. (a) Mean words across 13 tasks. (b) Short answers ($\leq$5 words) among FactSet-Homo responses covering the full reference (R-L=1). Target only generates LoRA from the relevant passage with D2L. Labels give counts.}
\label{fig:mode3}
\end{figure}

\textbf{Including the reference answer does not ensure the requested form.} We keep the document, question, and generated LoRA fixed, and change only the answering instruction. We examine responses that cover the entire reference answer, measured by R-L=1. Averaged over 13 tasks, only 4.5\% of these D2L responses have at most five words, compared with 75.7\% for ICL (Table~\ref{tab:prompt-response-control}). Even when D2L writes from only the passage containing the answer, short responses remain uncommon. Asking for an explanation instead of a short answer increases D2L's response length much less than ICL's (Figure~\ref{fig:mode3}). D2L on Gemma shows the same limited change in length (Table~\ref{tab:prompt-cross-instruction}).

\section{\framework{}}
\label{sec:method}
\begin{figure}[t]
\centering
\captionsetup{skip=4pt}
\includegraphics[width=\linewidth]{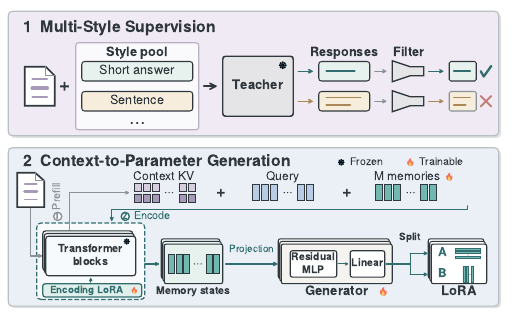}
\caption{\ours{} overview. Multi-style teacher answers supervise query-specific LoRA generation. Encoding LoRA acts on the query-and-memory suffix over a frozen document cache. Each layer's $M$ memory states generate a rank-$M$ update.}
\label{fig:memory-writer}
\end{figure}

The diagnoses motivate two design choices: using the query to guide which facts are written, and supervising how those facts are expressed. Given a document $D$ and query $q$, the full \framework{} model aggregates relevant content from a reusable document cache into learned memory states, then converts those states into LoRA updates. Multi-style teacher answers train the writer for factual recovery and response adaptation (Figure~\ref{fig:memory-writer}).

\subsection{Query-Guided Memory Construction}

A frozen copy of the answering backbone reads $D$ once with encoding LoRA disabled, storing its layerwise keys and values as $C(D)$. This cache is reused across queries about $D$. For each $q$, we process the query followed by $M$ learned memory tokens as a suffix attending to $C(D)$.

An \emph{encoding LoRA}, shared across documents and queries, adapts the attention and feed-forward projections for this suffix. Each memory token attends to the cached document, the query, and preceding memory tokens, combining document content according to the question. We denote its hidden state at layer $\ell$ and memory position $m$ by $h_{\ell,m}$. The default writer forms states per query without external relevance scores or attention bias.

Document states remain fixed during this process. New token projections and feed-forward computation are confined to the suffix, although suffix attention still accesses the full cache. Its cost therefore still grows with document length.

\paragraph{Training-free variant.}
We also introduce \ourstf{}, a training-free variant that applies query guidance to a pretrained writer. A frozen matcher scores document positions for the query. These scores bias the writer's extraction attention toward relevant content, and its existing generator converts the aggregated states into LoRA. Our main implementation uses released D2L with all writer parameters fixed. Appendix~\ref{app:method} details this variant and its extension to SHINE.

\subsection{Generating and Applying LoRA}
In the full model, we generate answering LoRA from the learned memory states at each layer $\ell$:
\begin{equation}
\begin{aligned}
 z_{\ell,m} &= P_\ell\operatorname{LN}(h_{\ell,m}),\\
 (A_\ell,B_\ell) &= G_\ell(z_{\ell,1:M}),\\
 \Delta W_\ell &= B_\ell A_\ell.
\end{aligned}
\end{equation}
Here $P_\ell$ projects each state into the generator's latent space. The generator $G_\ell$ uses a residual MLP, normalization, and a linear head to produce scaled LoRA factors. Each memory state supplies one rank component. We apply $\Delta W_\ell$ to the answerer's \texttt{down\_proj} weights at layer $\ell$.

The encoding LoRA is learned during training and reused to construct memory. The answering LoRA is generated for each document--query pair to supply information to the answerer.

\paragraph{Answering.}
The answerer receives the query and response instruction with the generated update applied. In standalone use, its decoding context contains neither the document nor the writer's memory tokens. The same update can also supplement document KV retained by SnapKV or DropKV (Section~\ref{sec:hybrid-analysis}). New documents and queries require only forward passes through the trained writer.

\subsection{Multi-Style Supervision}
To train the full model, we use the same frozen backbone as a teacher conditioned on the full document and query. It generates answers under varied instructions, including short answers, complete sentences, and explanations. The student uses the generated LoRA and the corresponding answering prompt, without the document. At each teacher-answer prefix, we match the student's next-token distribution to the teacher's using forward KL \citep{hinton2015distilling}. These targets teach content and answer form without passage labels.

\paragraph{Preserving base behavior.}
Generated LoRA changes the answerer's computation wherever the update is applied. We therefore combine answer distillation with a retention objective,
\begin{equation}
 \mathcal{L}=\mathcal{L}_{\mathrm{KL}}+\lambda_{\mathrm{retain}}\mathcal{L}_{\mathrm{retain}}.
\end{equation}
Here $\mathcal{L}_{\mathrm{KL}}$ is the tokenwise distillation loss above. On unrelated text, $\mathcal{L}_{\mathrm{retain}}$ matches predictions with the generated update to those of the unmodified backbone. It limits interference with base-model behavior, a concern also observed for document-trained LoRA \citep{eyuboglu2026cartridges}.

\paragraph{Training.}
We jointly train the memory embeddings, encoding LoRA, projections, and parameter generator. Gradients pass through the answerer to the generated factors and writer, while backbone weights remain frozen. Our default uses rank-32 encoding LoRA and $M=8$ memory tokens to generate rank-eight answering updates. Training runs for 2,000 steps with $\lambda_{\mathrm{retain}}=0.3$. Appendix~\ref{app:reproducibility} details data construction and training settings.

\raggedbottom
\clubpenalty=10000
\widowpenalty=10000
\section{Experiments}
\label{sec:experiments}

\subsection{Setup}
\textbf{Tasks and metrics.} We evaluate 17 tasks spanning FactSet, Wikipedia, SQuAD \citep{rajpurkar2016squad}, DROP \citep{dua2019drop}, NewsQA \citep{trischler2017newsqa}, TriviaQA \citep{joshi2017triviaqa}, Natural Questions \citep{kwiatkowski2019nq}, and LongBench \citep{bai2024longbench}, including HotpotQA \citep{yang2018hotpot}. We report token F1, exact match (EM), and ROUGE-L Recall (R-L) for reference coverage. Tasks receive equal weight.

\textbf{Baselines.} The Qwen3-4B comparison includes released D2L, further-trained D2L, and full-context ICL. We also evaluate \ourstf{}, our training-free variant using frozen D2L. The Mistral comparison adds OSCAR. Main results use the same short-answer instruction. SHINE is evaluated on its released Qwen3-8B backbone in the diagnoses and the \ourstf{} extension. Appendix~\ref{app:baselines} details the baseline settings.

\textbf{KV baselines and state budgets.} We compare with SnapKV, DropKV, Fast KVzip, and our reproductions of CCM and 500xCompressor (Appendix~\ref{app:kv-baselines}). Their budget equals 48 KV positions, matching \ours{}'s 6.75\,MiB output. Selected-KV baselines recompute the instruction and query over compressed document KV before answering. Budgets exclude prompt states, model weights, and construction caches.

\subsection{Comparison with Existing Methods}
\begin{table}[t]\centering
\caption{\looseness=-1 \tablelead{Comparison with parameter and KV baselines.} 17-task Qwen3-4B averages (Appendix~\ref{app:memory-token-results}). TF uses frozen D2L. KV methods and \ours{} use 6.75\,MiB. Teal marks best non-ICL scores.}
\label{tab:current-main}
\small
\setlength{\tabcolsep}{2.5pt}\renewcommand{\arraystretch}{1.15}
\setlength{\aboverulesep}{0pt}\setlength{\belowrulesep}{0pt}
\begin{tabular}{>{\raggedright\arraybackslash}p{\dimexpr.55\linewidth-2\tabcolsep\relax}*{3}{>{\centering\arraybackslash}p{\dimexpr.15\linewidth-2\tabcolsep\relax}}}\toprule
\rowcolor{PaperHeader}
Method & F1 & R-L & EM \\\midrule
ICL (full context) & 0.671 & 0.845 & 0.518 \\
\midrule
\rowcolor{PaperGroup}\multicolumn{4}{l}{\textit{Parameter writers}} \\
D2L & 0.173 & 0.514 & 0.016 \\
D2L + Further Training & 0.391 & 0.468 & 0.272 \\
\midrule
\rowcolor{PaperGroup}\multicolumn{4}{l}{\textit{KV baselines, 48 positions}} \\
SnapKV & 0.636 & \cellcolor{PaperBest}\textbf{0.789} & 0.485 \\
DropKV & 0.621 & 0.780 & 0.472 \\
Fast KVzip & 0.430 & 0.553 & 0.341 \\
500xCompressor & 0.520 & 0.606 & 0.392 \\
CCM & 0.485 & 0.559 & 0.358 \\
\midrule
\methodnametf{} & 0.260 & 0.755 & 0.025 \\
\methodname{} & \cellcolor{PaperBest}\textbf{0.663} & 0.774 & \cellcolor{PaperBest}\textbf{0.510} \\
\bottomrule\end{tabular}
\end{table}

\textbf{\ours{} improves on D2L baselines across different backbones.} On Qwen3-4B, the training-free variant improves over released D2L, while the full model exceeds both this variant and further-trained D2L on all metrics and approaches full-context ICL in F1 (Table~\ref{tab:current-main}). The advantage over released D2L also holds on Mistral and Gemma (Tables~\ref{tab:mistral-oscar} and~\ref{tab:gemma-detailed}). The training-free variant also improves Gemma D2L and Qwen3-8B SHINE on their respective tests (Table~\ref{tab:cross-writer}).

\begin{table}[t]\centering
\caption{\looseness=-1 \tablelead{Mistral-7B comparison.} Averages over 17 tasks, with best non-ICL scores highlighted. State measures document memory.}
\label{tab:mistral-oscar}
\small\setlength{\tabcolsep}{2.5pt}\renewcommand{\arraystretch}{1.15}
\setlength{\aboverulesep}{0pt}\setlength{\belowrulesep}{0pt}
\begin{tabular}{>{\raggedright\arraybackslash}p{\dimexpr.32\linewidth-2\tabcolsep\relax}>{\centering\arraybackslash}p{\dimexpr.26\linewidth-2\tabcolsep\relax}*{3}{>{\centering\arraybackslash}p{\dimexpr.14\linewidth-2\tabcolsep\relax}}}\toprule
\rowcolor{PaperHeader}
Method & State (MiB) & F1 & R-L & EM \\\midrule
ICL & 423.7 & 0.449 & 0.855 & 0.242 \\
\midrule
D2L & 9.0 & 0.294 & 0.535 & 0.125 \\
OSCAR & 32.0 & 0.218 & 0.762 & 0.010 \\
\midrule
\methodname{} & 9.0 & \cellcolor{PaperBest}\textbf{0.456} & \cellcolor{PaperBest}\textbf{0.780} & \cellcolor{PaperBest}\textbf{0.261} \\
\bottomrule\end{tabular}
\end{table}

\textbf{\ours{} is competitive with context compression.} At the same state budget on Qwen3-4B, \ours{} exceeds SnapKV, the strongest KV baseline in F1 and EM, although SnapKV and DropKV retain more reference content (Table~\ref{tab:current-main}). On Mistral, \ours{} also exceeds OSCAR's official checkpoint \citep{louis2025oscar} in F1 and EM with substantially less state (Table~\ref{tab:mistral-oscar}; setup in Appendix~\ref{app:mistral-oscar}). OSCAR has similar reference coverage but longer answers, so answer form also matters when interpreting its lower accuracy.

\subsection{Improvements on Diagnostic Tests}
\textbf{Fact discrimination.} On Wikipedia birth-year questions, \ours{} corrects nearly all of D2L's errors that borrowed another person's year. The remaining cases become other errors rather than correct answers (Appendix~\ref{app:targeted-repair}).

\textbf{Target retention.} With fixed target questions, \ours{} largely preserves accuracy as same-type facts are added, whereas D2L collapses. Other-type additions also preserve performance. Generic filler lowers F1 at similar coverage, showing that robustness depends on content (Table~\ref{tab:current-dilution-repair}).

\looseness=-1
\textbf{Response adaptation.} When asked for an explanation instead of a short answer, \ours{} expands its responses much more than D2L (Table~\ref{tab:prompt-seen}). It also more often answers correctly while satisfying held-out uppercase, JSON, and prefix formats (Table~\ref{tab:prompt-heldout-detail}). Response-length adaptation also holds on Gemma (Table~\ref{tab:prompt-cross-instruction}).

\subsection{Ablation Study}
\begin{table}[t]\centering
\caption{\tablelead{Writer, supervision, and memory-form ablations.} Qwen3-4B macro averages over the 17-task evaluation suite. Component comparisons share the training pool and step budget. Short-answer-only and D2L prompt are single-style controls; teacher recipes share documents, queries, encoder architecture, and step budget. Memory output forms share training data, training budget, and a 6.75\,MiB output size.}
\label{tab:current-components}
\small
\setlength{\tabcolsep}{2.5pt}\renewcommand{\arraystretch}{1.10}
\setlength{\aboverulesep}{0pt}\setlength{\belowrulesep}{0pt}
\begin{tabular}{>{\raggedright\arraybackslash}p{\dimexpr 0.55000000\linewidth-2\tabcolsep\relax}>{\centering\arraybackslash}p{\dimexpr 0.15000000\linewidth-2\tabcolsep\relax}>{\centering\arraybackslash}p{\dimexpr 0.15000000\linewidth-2\tabcolsep\relax}>{\centering\arraybackslash}p{\dimexpr 0.15000000\linewidth-2\tabcolsep\relax}}\toprule
\rowcolor{PaperHeader}
Configuration & F1 & R-L & EM \\\midrule
\rowcolor{PaperGroup}\multicolumn{4}{l}{\textit{(a) \ours{} writer variants}} \\
\methodname{} & 0.663 & 0.774 & 0.510 \\
w/o encoding LoRA & 0.562 & 0.650 & 0.405 \\
w/o query conditioning & 0.182 & 0.233 & 0.106 \\
w/ matcher & 0.665 & 0.775 & 0.511 \\
\addlinespace[4pt]
\rowcolor{PaperGroup}\multicolumn{4}{l}{\textit{(b) \ours{} teacher recipes}} \\
\methodname{} (multi-style) & 0.663 & 0.774 & 0.510 \\
Short-answer-only & 0.578 & 0.710 & 0.418 \\
D2L prompt & 0.307 & 0.777 & 0.088 \\
\addlinespace[4pt]
\rowcolor{PaperGroup}\multicolumn{4}{l}{\textit{(c) Memory output form}} \\
Generated LoRA & 0.663 & 0.774 & 0.510 \\
Memory Tokens KV & 0.682 & 0.807 & 0.522 \\
\bottomrule\end{tabular}
\end{table}

\textbf{Query guidance matters during writing.} Removing the query from construction sharply reduces accuracy, although the answerer still receives it. Removing encoding LoRA also hurts performance, while adding explicit matching to the complete encoder yields little improvement (Table~\ref{tab:current-components}a).

\textbf{Multi-style supervision improves F1 and EM over both single-style controls.} We compare multi-style supervision with short-answer-only and original-D2L-prompt supervision under matched training data, budget, and architecture. Multi-style training achieves higher F1 and EM than both controls and higher R-L than short-answer-only supervision (Table~\ref{tab:current-components}b), showing gains beyond using short-answer targets alone. The comparison with D2L-prompt also extends to varied answering instructions and held-out formats (Table~\ref{tab:teacher-format}).

\textbf{Generated LoRA largely preserves unrelated reasoning performance.} On the first 200 GSM8K questions, accuracy is 92\% with LoRA generated from unrelated documents, versus 94\% without LoRA under identical ICL prompts. Retention also reduces unrelated-text prediction drift with little change in QA accuracy (Appendix~\ref{app:retention-behavior}).

\textbf{Memory Tokens KV achieves higher standalone accuracy.} Memory Tokens KV directly uses the layerwise KV states of 48 learned memory tokens instead of generating LoRA. Training data, training budget, and output size are matched. Memory Tokens KV performs better on all metrics in standard standalone evaluation (Table~\ref{tab:current-components}c).

\begin{figure}[t]
\centering
\includegraphics[width=\linewidth]{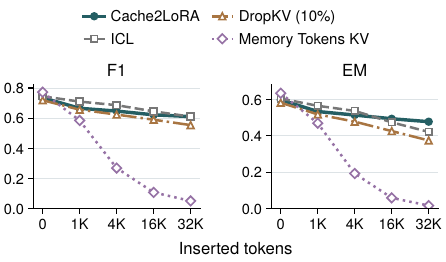}
\caption{Memory recall under intervening context. Macro F1 and EM over six tasks. Unrelated text precedes the question. LoRA, Memory Tokens KV, and selected document KV are fixed before insertion. LoRA and Memory Tokens KV each use 6.75\,MiB. ICL retains the full document. DropKV retains 10\% of document KV (minimum 48 positions).}
\label{fig:intervening-text}
\end{figure}

\textbf{LoRA supports more interference-resistant recall than Memory Tokens KV.} We insert up to 32K tokens of unrelated LongBench text before the answering question across six tasks. LoRA and Memory Tokens KV are constructed from the original document and query, then fixed across lengths. ICL retains the full document, while DropKV retains a subset of document KV. At the longest insertion, \ours{} has higher macro EM than all three alternatives and similar F1 to ICL (Figure~\ref{fig:intervening-text}). Memory Tokens KV declines sharply, falling below DropKV once text is inserted. Its reference coverage also drops sharply, indicating that the accuracy loss is not merely a consequence of longer answers. Memory Tokens KV relies on attention to a fixed memory prefix, whose use may be disrupted by intervening text \citep{shi2023irrelevant}. Its use may also be position-sensitive \citep{liu2023lost}. Shifting memory-key RoPE positions to restore their original distance to the question substantially improves KV recall at 4K. At 32K, however, it remains far below LoRA on all six tasks despite this correction (Table~\ref{tab:intervening-position}), so this positional mismatch alone does not explain the gap. At 4K, aligned KV even exceeds LoRA in F1 on two tasks, showing that positional handling materially affects the comparison. The persistent gap at 32K distinguishes initial answer recovery from continued use of a constructed memory: generated LoRA is more robust in the latter setting, even though learned KV performs better in standard standalone evaluation (Table~\ref{tab:current-components}c). LoRA modifies computation at each subsequent position without requiring memory-prefix retrieval, potentially supporting more stable recall.

\subsection{Hybrid Answering}
\label{sec:hybrid-analysis}
\begin{figure}[t]
\centering
\includegraphics[width=\linewidth]{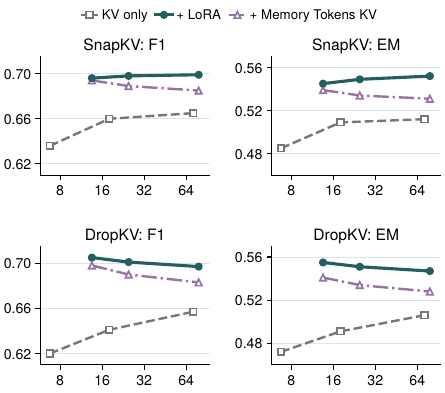}
\caption{Hybrid F1 and EM with SnapKV and DropKV. The horizontal axis gives total document-derived state (MiB, log scale). Generated LoRA and Memory Tokens KV each add 6.75\,MiB to the same selected document cache at $B=48,128,512$.}
\label{fig:decoding-state-quality}
\end{figure}

Prior work combines document-trained LoRA with compressed KV \citep{zuo2026rethinking}. We test generated updates with document positions selected by SnapKV or DropKV. Both selectors use the query. All conditions process the instruction and query again over the selected cache, including controls without LoRA. The writer and selectors are combined without joint training (Appendix~\ref{app:kv-protocol}).

\textbf{LoRA combines better with selected KV on F1 and EM.} We replace LoRA with Memory Tokens KV while keeping the selected document cache fixed. Both add 6.75\,MiB. Across all tested budgets, LoRA gives higher F1 and EM with both SnapKV and DropKV (Figure~\ref{fig:decoding-state-quality}). Memory Tokens KV has higher R-L at all three budgets.

\textbf{The update's source and query matter.} Holding the selected cache and answering procedure fixed, enabling \ours{} improves F1 with both selectors. Replacing it with an update from another document lowers performance below KV-only answering. An update for a different query about the same document also underperforms KV-only answering on the matched subset (Table~\ref{tab:hybrid-controls}). The gains thus depend on an update aligned with both the source and the current question, rather than merely adding an adapter. In particular, the same-document control holds the source information fixed while changing which question guides memory construction.

\textbf{LoRA helps even when answer positions are retained.} When selection prioritizes positions containing the reference answer, adding \ours{} still improves F1 and EM with both selectors (Table~\ref{tab:hybrid-retention}). Its benefit is therefore not confined to cases where compression removes those positions. This supports a complementary role in using retained evidence, although retaining answer positions does not guarantee that all supporting context survives. With conflicting facts, Wikipedia answers often follow LoRA under tighter compression but shift toward KV as more positions are retained (Table~\ref{tab:hybrid-conflict}).

These results support using generated LoRA as a complement to retained document KV. The gains persist across both selectors and all tested budgets, up to 512 retained positions, without joint training, indicating that the update remains useful as more source context is retained.

\subsection{Efficiency}
\begin{figure}[t]
\centering
\includegraphics[width=\linewidth]{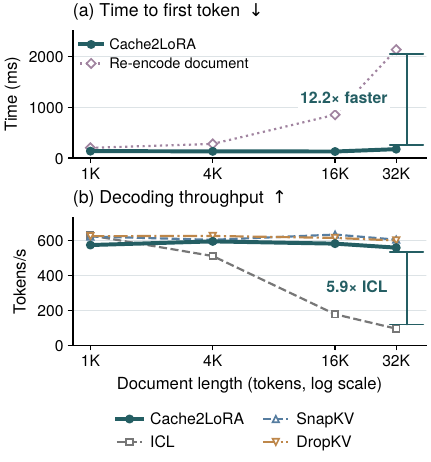}
\caption{Qwen3-4B efficiency on one H200. (a) Time from the query to the first answer token, excluding one-time document caching. \ours{} generates and applies LoRA, while the control re-encodes the document for each query. (b) Batch-16 decoding throughput with 256 output tokens. SnapKV and DropKV retain 48 positions. Full results are in Table~\ref{tab:efficiency-cached-full}b.}
\label{fig:efficiency-scaling}
\end{figure}

On one H200, we measure one-time document caching, time to the first answer token after caching, and batch-16 decoding with 256 output tokens.

\looseness=-1
\textbf{Caching benefits long documents.} Relative to re-encoding on the same backbone, \ours{} reaches the first answer token 2.1$\times$ faster at 4K, 6.6$\times$ at 16K, and 12.2$\times$ at 32K (Figure~\ref{fig:efficiency-scaling}a). The control re-encodes the document and suffix for each question. These ratios exclude one-time document caching, which costs 1,820.6\,ms for \ours{} at 32K and is shared across queries (Table~\ref{tab:efficiency-cached-full}b). At 32K, cached 16-token answering is 4.6$\times$ faster than the re-encoding control (Table~\ref{tab:efficiency-output-length}c).

\textbf{Compact memory benefits long-document batch decoding.} As document length grows from 4K to 32K, \ours{} throughput changes from 596 to 560 tokens/s, while full-context ICL drops from 512 to 95 tokens/s. The resulting speedup grows from 1.2$\times$ to 5.9$\times$ (Figure~\ref{fig:efficiency-scaling}b). At 32K, per-request decoding state is 12.52\,MiB versus ICL's 4,613.91\,MiB, excluding the separately retained document cache. SnapKV and DropKV remain faster at 606 and 601 tokens/s with the current inference infrastructure. Merging LoRA into the weights avoids memory-prefix attention, but our measurements execute adapter factors separately for each request. This batch-throughput advantage does not imply lower single-request latency: cached ICL is faster in the tested 32K settings with 16- and 256-token outputs (Table~\ref{tab:efficiency-output-length}c). Table~\ref{tab:efficiency} reports all state and timing results.

\section{Conclusion}
We diagnose fact discrimination, target retention, and response adaptation in document-to-LoRA writers. Source interventions and fixed-capacity controls motivate query-guided information aggregation, while instruction probes motivate supervision of answer expression. \framework{} combines these choices with reusable document caching, improving on released D2L, \ourstf{}, and further-trained D2L. Its generated updates also supplement selected document KV, with gains persisting at larger retained-context budgets.

\looseness=-1
At matched memory size, generated LoRA shows two strengths relative to the learned-KV variant: it preserves answer accuracy more reliably under long, diverse intervening text and complements retained document KV more effectively, yielding higher hybrid F1 and EM across both selectors and all tested budgets. These results support generated parameter memory for interference-resistant recall and hybrid answering.

\clearpage
\section*{Limitations}

\framework{} constructs a new LoRA update for each query. The full document cache is needed during construction, while decoding uses the compact output. The cache can be stored on CPU between construction calls, though our measurements use GPU-resident caches and do not include transfer latency. Per-query construction adds preparation time and transient GPU memory. We report preparation latency, decoding-state size, and the document-prefill peak allocation increment (Appendix~\ref{app:efficiency-measured}), rather than a total deployment peak including all resident models and caches. Our experiments assume fixed source documents. Maintaining caches and constructed memories as documents are revised remains future work.

The current evaluation considers individual queries. In multi-turn interactions, relevant context can change as a conversation develops, creating choices about when to retain, refresh, or replace a constructed memory. We leave the design and evaluation of these memory management policies for conversational use to future work.

Finally, the writer constructs each update from the context available for a query. Our hybrids combine this output with selected document KV. Composing memories constructed independently from separate documents is a distinct setting that we have not evaluated. Such composition requires deciding which memories to combine and how to reconcile conflicting information across them.

\section*{Ethical Considerations}
This work studies document memory using existing public benchmarks, synthetic facts, and pretrained language models. Their use remains subject to the original licences and access conditions. Compression into LoRA or KV states should not be treated as anonymization: either form may preserve sensitive source information and reproduce it in answers. Applications should use authorized documents, control access to derived memories, and account for source deletion or correction. Generated answers may also retain errors or biases from the source text and backbone.

Generative AI tools assisted with analysis code and manuscript editing. The authors remain responsible for the methods, results, and final text.

\bibliography{refs}

\clearpage
\appendix
\setcounter{dbltopnumber}{4}
\setcounter{topnumber}{4}
\setcounter{totalnumber}{6}
\renewcommand{\dbltopfraction}{0.95}
\renewcommand{\topfraction}{0.95}
\renewcommand{\textfraction}{0.05}
\renewcommand{\dblfloatpagefraction}{0.75}

\raggedbottom
\twocolumn[{
\begin{minipage}{\textwidth}
\section{Experimental Setup}
\label{app:reproducibility}
\subsection{Training Data}
\begin{center}
\centering
\small
\setlength{\tabcolsep}{5pt}
\renewcommand{\arraystretch}{1.12}
\begin{tabular}{lrrrrr}
\toprule
\rowcolor{PaperHeader}
Source & \multicolumn{2}{c}{Before teacher filtering} & \multicolumn{2}{c}{After teacher filtering} & Mix (\%) \\
\rowcolor{PaperHeader}
& Documents & Queries & Documents & Queries & \\
\midrule
SQuAD & 5,000 & 23,522 & 5,000 & 22,902 & 18.67 \\
DROP & 3,000 & 21,456 & 2,988 & 17,707 & 11.12 \\
ROPES & 932 & 6,119 & 932 & 5,877 & 3.48 \\
PwC & 3,000 & 23,485 & 3,000 & 22,804 & 11.21 \\
\addlinespace[3pt]
SQuAD, same-article groups & 2,000 & 16,000 & 2,000 & 15,540 & 7.47 \\
SQuAD, same-type answers & 2,439 & 14,037 & 2,439 & 13,652 & 9.11 \\
SQuAD, long concatenations & 2,127 & 16,611 & 2,127 & 16,092 & 7.95 \\
DROP, long concatenations & 827 & 6,522 & 827 & 4,968 & 3.09 \\
\addlinespace[3pt]
Synthetic, competing facts & 1,500 & 9,024 & 1,500 & 9,024 & 5.60 \\
Synthetic, shared subjects & 1,500 & 8,956 & 1,500 & 8,946 & 5.60 \\
Synthetic biographies & 1,969 & 11,077 & 1,969 & 11,069 & 7.36 \\
Synthetic, long multi-hop & 2,500 & 15,028 & 2,500 & 12,771 & 9.34 \\
\midrule
Total & 26,794 & 171,837 & 26,782 & 161,352 & 100.00 \\
\bottomrule
\end{tabular}
\captionof{table}{\tablelead{Training data for the main Qwen3-4B model.} Documents and queries count source records before style expansion. After filtering, a record is counted if at least one response survives. Mix is the source's share of retained document variants in the training pool.}
\label{tab:training-data}
\end{center}

\end{minipage}
\vspace{1em}
}]
\subsubsection{Sources and Input Construction}
Training uses the training splits of SQuAD, DROP, ROPES, and PwC, together with synthetic facts. Natural inputs group questions by document. HotpotQA, 2WikiMultihopQA, and MuSiQue are excluded from the training corpus. Table~\ref{tab:training-data} reports source documents and queries before and after teacher-response filtering; multiple teacher responses do not count as additional source records.

\paragraph{Document construction and separation.}
\looseness=-1
We combine single passages with multi-passage inputs containing related facts or distractors. Synthetic documents include competing facts, shared subjects, biographies, and relational questions. Training inputs are limited to 3,072 tokens. Before teacher generation, overlap checks exclude matches to evaluation and calibration material using passage overlap and query similarity. Synthetic names and templates are also separated from held-out pools. Long natural documents combine four to six source units, with at most two questions per unit and up to eight questions per document, plus shuffled filler passages, targeting 1,500--3,072 tokens. Long synthetic documents contain 16--26 fictional people and four to eight one- or two-hop questions. Construction uses seed 1009. A separate calibration set monitors training, and all reported models use the final checkpoint. Appendix~\ref{app:setup} reports the overlap audit.

\paragraph{Overlap thresholds.}
For overlap checks, text is lowercased and reduced to sequences of word characters. A passage is excluded if at least 10\% of its word 13-grams, or at least 50 such occurrences, appear in the evaluation/calibration exclusion index. Exact matches to indexed short passages or sentences of at least six words are also excluded, along with matched source titles where available. A question is excluded when its normalized text matches exactly, or its word-trigram Jaccard similarity is at least 0.6, and an accepted-answer pair has token F1 of at least 0.5. These checks are repeated on the constructed documents. Synthetic question templates are additionally compared after replacing capitalized names by a placeholder and digit sequences by zero. Matching templates are removed, and synthetic documents with fewer than three remaining questions are discarded. Source passages may recur across training constructions, but long natural-document variants use previously unused normalized questions. The two style variants of each document are retained intentionally.

\subsubsection{Teacher Responses and Sampling}
\paragraph{Teacher prompting.}
A frozen teacher produces responses in eight styles: evaluation short answer, original D2L, direct answer, explicit short answer, entity or number, complete sentence, extractive span, and brief explanation. Before compatibility filtering, their sampling probabilities are 0.20, 0.15, 0.15, 0.10, 0.12, 0.12, 0.10, and 0.06, respectively. For each query, two compatible styles are sampled sequentially without replacement, renormalizing these weights over the remaining eligible styles. Extractive styles require a verbatim accepted-answer span, excluding the base DROP and ROPES sources and yes/no answers. Entity styles require an accepted answer of at most five words and exclude yes/no, why, and general how questions. Count questions additionally require an answer span in the document. PwC uses direct-answer and D2L styles. The single-style controls use either short-answer-only or original-D2L responses, with the same sampled training data and training budget as the multi-style model. During distillation, the student answers without the document using the corresponding response style.

\paragraph{Filtering and training mixture.}
Teacher generation is greedy with a 256-token limit. We remove empty, truncated, or unsuitable responses and filter short-answer targets against accepted answers; PwC is exempt from the short-answer correctness filter. After normalizing thousands separators and English number words from zero to twenty, a response is removed if it has zero token F1 against all accepted answers and contains none of them. Refusals without an accepted-answer match and queries shorter than three characters are also removed. Documents with no retained questions are discarded. Retained variants are pooled without additional source weighting and reused throughout training. The D2L-style control uses the same construction pool and filtering rules. Gemma generates its own teacher responses from this pool and applies the input limit with its own tokenizer.

Table~\ref{tab:teacher-prompts} gives the teacher templates. Table~\ref{tab:teacher-wrappers} lists the 22 closed-QA wrappers used in the three legacy styles. For the three legacy styles, the student receives the wrapped question alone. For the remaining styles, it receives the same instruction and raw question with the document prefix removed.
\begin{table*}[t]
\centering\small
\caption{\tablelead{Teacher prompts and sampling weights.} Each template is a single user message in the backbone's chat template, followed by the assistant-generation prefix. $D$ denotes the document. For the original-D2L, direct-answer, and explicit-short-answer styles, $q$ includes a sampled D2L closed-QA instruction, except for PwC. Other styles use the raw question. Table~\ref{tab:teacher-wrappers} lists the closed-QA wrappers. Blank lines are condensed here.}
\label{tab:teacher-prompts}
\setlength{\tabcolsep}{4pt}
\begin{tabular}{>{\raggedright\arraybackslash}p{.16\textwidth}>{\raggedright\arraybackslash}p{.06\textwidth}>{\raggedright\arraybackslash}p{.72\textwidth}}
\toprule
\rowcolor{PaperHeader} Style & Weight & User-message template \\
\midrule
Evaluation short answer & 0.20 & $D$\newline Answer the following question. Output only the answer and do not output any other words.\newline Question: $q$ \\ \addlinespace[4pt]
Original D2L & 0.15 & You are an honest and helpful assistant.\newline \# Provided Information\newline $D$\newline ---\newline \# System Instruction\newline - The information provided is up-to-date information and/or the user instruction.\newline - When the provided information is not relevant to the question, \textbf{ignore} it and answer the question based on your knowledge.\newline - If the provided information is related to the question, incorporate it in your response.\newline - If the provided information is an instruction, follow the instruction carefully.\newline ---\newline \# User Input\newline $q$ \\ \addlinespace[4pt]
Direct answer & 0.15 & Document:\newline $D$\newline Answer the question directly.\newline $q$ \\ \addlinespace[4pt]
Explicit short answer & 0.10 & Document:\newline $D$\newline Answer the question directly. Output only the answer and do not output any other words.\newline $q$ \\ \addlinespace[4pt]
Entity or number & 0.12 & Document:\newline $D$\newline Answer with only the name, entity, date or number that answers the question. Do not write a sentence.\newline Question: $q$ \\ \addlinespace[4pt]
Complete sentence & 0.12 & Document:\newline $D$\newline Answer the question in one short, complete sentence.\newline Question: $q$ \\ \addlinespace[4pt]
Extractive span & 0.10 & Document:\newline $D$\newline Copy the exact words from the document that answer the question. Output only that span, verbatim.\newline Question: $q$ \\ \addlinespace[4pt]
Brief explanation & 0.06 & $D$\newline Answer the following question, then explain your answer in one or two sentences.\newline Question: $q$ \\ \addlinespace[4pt]
\bottomrule
\end{tabular}
\end{table*}

\begin{table*}[t]
\centering\small
\caption{\tablelead{D2L closed-QA wrappers.} One of these 22 templates is sampled uniformly for the original-D2L, direct-answer, and explicit-short-answer styles. $q_0$ is the raw question. The resulting string supplies $q$ in Table~\ref{tab:teacher-prompts} and serves as the student's prompt. PwC uses its original query.}
\label{tab:teacher-wrappers}
\setlength{\tabcolsep}{4pt}
\begin{tabular}{rp{.91\textwidth}}
\toprule
\rowcolor{PaperHeader} No. & Template \\
\midrule
1 & Answer the question based on the given passages. Only give me the answer and do not output any other words.\newline Question: $q_0$ \\ \addlinespace[2pt]
2 & Answer without any explanation.\newline Question: $q_0$ \\ \addlinespace[2pt]
3 & Based on the provided text, what is the answer to the following question? Provide only the answer.\newline Question: $q_0$ \\ \addlinespace[2pt]
4 & Extract the answer to the question from the text. Be concise. Do not explain.\newline Question: $q_0$ \\ \addlinespace[2pt]
5 & What is the answer to this question, based on the context? Respond with the answer only.\newline Question: $q_0$ \\ \addlinespace[2pt]
6 & Provide a direct answer to the question using the given passages. Do not give any explanation.\newline Question: $q_0$ \\ \addlinespace[2pt]
7 & Answer the question using only information from the provided text. No extra words.\newline Question: $q_0$ \\ \addlinespace[2pt]
8 & From the passages, answer the question. Just the answer, please.\newline Question: $q_0$ \\ \addlinespace[2pt]
9 & Give the answer to the question. Do not include any other text.\newline Question: $q_0$ \\ \addlinespace[2pt]
10 & The answer to the question is in the text. Find it and state it clearly. No need for explanation.\newline Question: $q_0$ \\ \addlinespace[2pt]
11 & Concisely answer the question based on the text provided. Don't include any other words. Just the answer.\newline Question: $q_0$ \\ \addlinespace[2pt]
12 & Read the passages and answer the question with the minimal necessary words.\newline Question: $q_0$ \\ \addlinespace[2pt]
13 & What is the direct response to the question, according to the text? Avoid explanation.\newline Question: $q_0$ \\ \addlinespace[2pt]
14 & Please provide only the answer to the question, derived from the text.\newline Question: $q_0$ \\ \addlinespace[2pt]
15 & Using the provided context, answer the question. Output the answer and nothing else.\newline Question: $q_0$ \\ \addlinespace[2pt]
16 & Identify the answer in the text and present it without elaboration.\newline Question: $q_0$ \\ \addlinespace[2pt]
17 & Answer the following question based on the text. Your answer should be brief and to the point. No explanation.\newline Question: $q_0$ \\ \addlinespace[2pt]
18 & Based on the information given, what is the answer to the question? Only state the answer.\newline Question: $q_0$ \\ \addlinespace[2pt]
19 & Find the answer to the question in the provided passages and write it down. No explanations.\newline Question: $q_0$ \\ \addlinespace[2pt]
20 & The question is: $q_0$. Provide the answer based on the text, and nothing more. \\ \addlinespace[2pt]
21 & Question: $q_0$\newline Answer directly based on the text provided. No extra words. \\ \addlinespace[2pt]
22 & Question: $q_0$\newline Please provide the answer based on the text. No explanation is needed. \\ \addlinespace[2pt]
\bottomrule
\end{tabular}
\end{table*}

\subsection{Evaluation and Scoring}
\label{app:setup}
\paragraph{Datasets and cohorts.}
The main benchmark contains 17 tasks. FactSet and Wikipedia tasks use 304 questions each, document QA and constructed multi-document tasks use 300, and LongBench tasks use 200 (150 for MultiFieldQA-en). The duplicate non-LongBench HotpotQA task is excluded. Cross-writer comparisons use the common thirteen-task cohort without LongBench. Comparisons align questions and accepted answers, and task averages receive equal weight even when related tasks are grouped for display. Qwen evaluations use complete benchmark inputs.

\paragraph{Concatenated inputs.}
SQuAD-4 combines an answer paragraph with three others from the same article, SQuAD-Mix groups shuffled paragraphs, and NQ-4 combines answer passages from different queries. Each question receives a separate update constructed from the concatenated input.

\paragraph{Answering instructions.}
\looseness=-1
Unless varied explicitly, comparisons use the D2L short-answer instruction: ``Answer the following question. Output only the answer and do not output any other words.'' Generation is greedy with a 256-token limit. Response-adaptation tests instead request an entity or number, one complete sentence, a brief explanation, or use the question alone. Documents, factual questions, and generated memory remain fixed while the answering instruction changes. The writer receives the raw question, without the varying response instruction. Qwen sweeps cover the main tasks. Cross-method averages cover thirteen shared tasks. Qwen/Gemma comparisons use three untruncated tasks.

\paragraph{Metrics.}
We report normalized token F1, exact match (EM), and ROUGE-L Recall (R-L). R-L measures reference coverage, while containment credits an accepted answer appearing anywhere in the response. Scores take the best accepted reference and are averaged equally across tasks. Short responses contain at most five words. Conditional brevity is measured among responses with full reference coverage, excluding tasks without such responses. Longer answers are scored against the same short-answer references, so their accuracy is interpreted alongside length and coverage.

Response-length span is the difference between the longest and shortest task-average responses across instructions, divided by the corresponding ICL span. Normalization matches backbones, tasks, and instructions: four on Qwen, and short answer plus explanation on Gemma.

\paragraph{Held-out formats.}
Uppercase, JSON, and an \texttt{ANSWER:} prefix are absent from the teacher styles. JSON and prefix tests use SQuAD-4, NQ-4, Wikipedia birth year, and NQ pages. Uppercase evaluation uses the same subset for every method, restricted to references containing cased letters and therefore excluding numeric birth years. Format success requires uppercase text, a bare JSON object containing only a string-valued answer, or the prescribed prefix, respectively. Content scores use the extracted answer; joint success requires both correct formatting and exact-match correctness. For separate content scoring, removable formatting wrappers do not count as answer content.

\paragraph{Training--evaluation overlap.}
An exact-paragraph audit after normalization finds no training passages in the benchmark. Query matches occur in DROP and MultiFieldQA-en, but their passages differ. SQuAD and DROP share training source families, but not tested passages.

\subsection{Model Architecture and Training}
\label{app:new-method}
\paragraph{Architecture.}
\ours{} uses a frozen Qwen3-4B document encoder with bfloat16 computation. Rank-32 encoding LoRA processes the query and eight learned memory tokens over the unchanged document cache. A trainable projection and residual generator convert their states into a rank-eight answering update for the feed-forward output projections. The generator is initialized from D2L and trained jointly with the encoder. The encoding LoRA uses scaling $\alpha/r=32/32=1$, with zero dropout, on the attention query, key, value, and output projections and all three feed-forward projections. Generator dimensions and output scaling are specified in Appendix~\ref{app:generator-details}. The default omits explicit matching and D2L's static bias adapter.

\paragraph{Training.}
Training uses forward KL distillation with a retention weight of 0.3 for 2,000 steps, with 32 document variants per step. Encoder and generator learning rates are $10^{-4}$ and $2\times10^{-6}$, with 100 warmup steps. We use AdamW with $(\beta_1,\beta_2)=(0.9,0.999)$, $\epsilon=10^{-8}$, zero weight decay, gradient-norm clipping at 1.0, and seed 0. After linear warmup, cosine decay reduces each learning rate to 10\% of its peak. Each document contributes at most eight retained question--response pairs. The backbone remains frozen. Component controls share these defaults unless noted.

\paragraph{Ablation controls.}
Removing encoding LoRA leaves the backbone encoder unadapted while training the memory embeddings and generator. Removing query conditioning hides the query during writing in both training and evaluation; the answerer still receives it. The explicit-matching control adds relevance biases to the trained writer, whereas \ourstf{} keeps the writer frozen. Memory Tokens KV retains the query-conditioned encoder but directly uses the layerwise KV states of 48 learned memory tokens instead of generating LoRA. It matches LoRA in output size (6.75\,MiB), multi-style training data, and schedule.

We use two single-style supervision controls: short-answer-only targets and original-D2L teacher responses (D2L-prompt). Both sample the same training data and use the same training budget as the multi-style model. Architecture, objective, and optimization settings are held fixed. A separate D2L-prompt control omits retention.

\paragraph{Other backbones.}
Gemma-2-2B is trained separately using its own teacher and the same training recipe. Its generator is initialized from the corresponding D2L checkpoint. To accommodate the evaluation cache, ten NQ-page documents are truncated to 4,032 tokens; other inputs in the thirteen-task cohort are untruncated. The Mistral model uses Mistral-7B-Instruct-v0.2 teacher responses and the same multi-style schedule.

\subsection{Baselines}
\label{app:baselines}
Base answers without document information, while ICL receives the full document. Released D2L and SHINE checkpoints provide the parameter-writer baselines, with their native parameter scaling and document-independent components preserved \citep{charakorn2026d2l,liu2026shine}. Their writers do not receive the answering query. See Appendix~\ref{app:method} for the training-free variant.

\paragraph{D2L + Further Training.}
We continue training the released D2L checkpoint with its baseline recipe, retaining the query-independent writer and static bias adapter. The hypernetwork is trained while the backbone remains frozen, using D2L-prompt forward-KL distillation and retention with weight 0.3. The hypernetwork learning rate is $2\times10^{-6}$ with 50 warmup steps.

\paragraph{Learned KV baselines.}
\label{app:kv-baselines}
We reproduce CCM and 500xCompressor on Qwen3-4B using our multi-style data and training schedule. CCM applies encoding LoRA only to appended memory positions, whereas 500xCompressor also adapts document encoding. Both produce 48-position KV without receiving the query during construction; the answerer receives the query over this prefix. These are adaptations to our backbone and training recipe, rather than the original released training runs.

\paragraph{Selected KV.}
SnapKV and DropKV select document positions using the query \citep{li2024snapkv,zhang2026dropkv}. Fast KVzip uses released gating weights with global allocation \citep{kim2026fastkvzip}. Standalone comparisons match the state budget of \ours{}; additional budget comparisons appear in Appendix~\ref{app:kv-results}.

\paragraph{OSCAR.}
\label{app:oscar-protocol}
We use the official \texttt{naver/oscar-mistral-7B} checkpoint, including its trained answerer. OSCAR compresses document segments jointly with each query using its native segment and memory configuration. It receives the same short-answer instruction and generation limit as the other methods. Its state budget is computed from the resulting memory prefix and averaged across tasks.

\subsection{Diagnostic Protocols}
\label{app:diagnostic-protocols}
The probes hold target questions fixed while changing source content, construction, or answering instructions. Target-only controls use annotated supporting evidence; directly optimized reference adapters preserve the size and structure of generated updates. Internal probes use D2L on Qwen3-4B unless stated otherwise.

\paragraph{Fact discrimination and repair.}
Wikipedia probes add competing people or replace their facts. Birth-year responses are classified by the first year as correct, borrowed from another person, or other, with correctness taking precedence when years coincide. Source interventions alter the borrowed source or average its memory values, using another source as a control. The intervention cohort contains 192 questions; the repair cohort contains 304. Repair counts track the same baseline-borrowed cases, and gold labels are used only for evaluation.

\paragraph{Target retention.}
Synthetic probes hold 64 target questions fixed while adding same-type facts, other-type facts, or empty content. Target amplification restores the target's isolated attention share without renormalizing or directly suppressing distractors. Value averaging replaces distractor values by their mean while retaining their keys; gold-support bias favors annotated evidence. A natural-text probe adds distractor passages or generic filler to 60 fixed SQuAD targets. Filler avoids names, dates, and numbers. Attention enrichment measures target attention relative to its share of input tokens.

\paragraph{Query dependence and attention sharing.}
For a fixed adapter, the activation probe measures how much its direct contribution varies across questions, normalized by total activation energy. It uses eight documents with eight questions each and measures the direct document-dependent contribution, excluding the shared bias from the measured vector. Appendix~\ref{app:aliasing-internals} defines the statistic. Attention sharing is measured by the overlap between facts' combined extraction and assembly weights, averaged across layers on sixteen FactSet-Homo documents.

\paragraph{KV and capacity controls.}
Learned KV compressors undergo the same source-count, fixed-target expansion, and response-instruction probes, with matched questions within each comparison. Target-only controls use annotated evidence. Capacity controls optimize the writer's internal memory states while freezing the backbone and downstream parameter mapping. Each resulting adapter is shared across all questions about its document. D2L uses 200 distillation steps with retention; its learning rate is calibrated on the first document, which remains in the reported cohort. SHINE uses the analogous frozen-mapping intervention.

\subsection{KV Compression and Hybrid Protocols}
\label{app:kv-protocol}
All methods use matched documents, queries, instructions, and scoring. SnapKV and DropKV retain only document positions using their official selection procedures. The instruction and query are then processed again over the selected cache, and the first answer token is generated from this recomputation. No full-document query states or first-token prediction are reused. Selected keys retain their original positional encoding.

\paragraph{Hybrid construction.}
Hybrid comparisons match unique document--query pairs across conditions, removing repeated records. Standalone benchmark tables retain the original evaluation records.

We combine the selected cache with either generated LoRA or Memory Tokens KV without joint training, sweeping 48, 128, and 512 retained document positions. Both additions use 6.75\,MiB. Learned memory KV is appended after the selected document cache, with answering positions continuing after the document and memory prefixes. Selected positions and answering are identical with and without adapters.

Fast KVzip uses default global allocation across layers and heads, with a matched total budget and the same answering procedure.

\paragraph{Evidence interventions.}
Adapter substitutions use an update from another document or another question about the same document while holding the selected cache fixed. Answer-retention controls prioritize positions containing an accepted answer and their immediate neighbors, filling the remaining budget by the original selection scores. Inputs without an answer match use ordinary selection. Counterfactual tests combine caches and adapters from original and altered documents while keeping the question fixed; containment of the original and altered answers is scored separately.

\subsection{Efficiency Measurement and State Accounting}
Measurements use one H200, Qwen3-4B, and a shared HuggingFace implementation, with a 4-bit constructor and bfloat16 answerer. Document caching is paid once; per-query preparation includes constructing and applying memory through the first answer token. Decoding tests vary document and output length, with batch-one latency and batch-16 throughput reported in Table~\ref{tab:efficiency}. Timings use repeated measurements after warmup with GPU synchronization, excluding tokenization and CPU transfers. Batched LoRA executes separate adapter factors for each request; timings exclude online scheduling and shared-prefix paging.

LoRA construction uses eight learned memory tokens to generate a rank-eight update; the Memory Tokens KV variant uses 48 learned memory tokens as its KV output. Both outputs occupy 6.75\,MiB. The re-encoding control processes the document, query, and memory-token suffix for every question on the same backbone. It measures that computational path, not the accuracy or implementation of another published compressor.

Document-derived state counts retained document KV and generated adapters. Total decoding state additionally includes prompt states. Both exclude shared model weights, document caches retained for future queries, padding, and allocator overhead. The rank-eight update and 48-position KV each occupy 6.75\,MiB.

\subsection{Intervening-Text Protocol}
\label{app:intervening-protocol}
\paragraph{Fixed memory and shared inputs.}
We test 1,808 questions from SQuAD-4, NQ-4, Wikipedia birth year, FactSet-Homo, NewsQA, and SQuAD-Mix. LoRA and Memory Tokens KV are constructed once from the original document and query, then held fixed across insertion lengths. Both use the checkpoints from the main comparisons and occupy 6.75\,MiB. DropKV selects 10\% of document positions using the original query, with a minimum of 48 or the full document if shorter, and also keeps this selection fixed. ICL retains the full document.

\looseness=-1
All methods receive identical unrelated text before the D2L short-answer instruction and question. Passages come from the English LongBench HotpotQA, 2Wiki, MuSiQue, and MultiFieldQA subsets, excluding duplicates and passages containing accepted answers. Longer conditions extend the same shuffled passage stream, up to 32K inserted tokens. Answers use greedy short-answer decoding.

\paragraph{Memory-key RoPE alignment.}
\label{app:position-control}
Construction is unchanged: the writer processes the document prefix, then the query and 48 memory tokens. Their cached keys carry RoPE positions $o-48,\ldots,o-1$, where $o$ is the total construction length. Answering still processes the template, intervening text, and question over the memory prefix, with prompt position IDs starting at $o$. The alignment control leaves the input sequence, prompt position IDs, and attention mask unchanged, retaining all intervening tokens. It shifts only memory-key positions by the added prompt length $L$, keeping values fixed:
\begin{equation}
 k' = k\odot\cos(L\theta)
       + \operatorname{rotate\_half}(k)\odot\sin(L\theta).
\end{equation}
The shift restores the memory-to-question RoPE offsets of the no-insertion condition. We use the answerer's rotary embedding and Qwen3 rotation convention, compute in float32, and cast back to bfloat16. Both writer and answerer use RoPE base 5,000,000 without scaling.

At $L=0$, keys are bitwise unchanged and all 300 SQuAD-4 predictions match the original variant. In an eight-case check, shifting both memory keys and prompt positions by 4,096 or 32,768 preserves all first-token argmax predictions, with maximum logit differences of 0.25--0.31. Shifting only keys changes these predictions in four to eight cases, with maximum logit differences of 13--19. These checks confirm that the intervention changes positional alignment while preserving the zero-shift behavior.

\section{Additional Diagnostic Evidence}

\subsection{Fact Discrimination: Responses and Source Attribution}
\label{app:aliasing-internals}
\begin{figure}[!tb]\centering
\includegraphics[width=\linewidth]{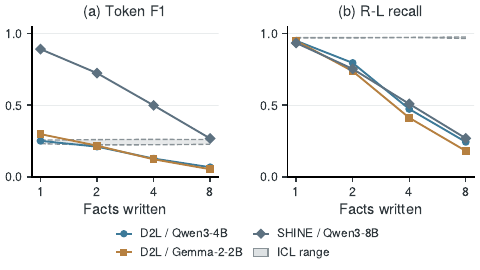}
\caption{Wikipedia fact-count probe: 304 matched queries per count, using raw queries. Gray bands span the minimum and maximum ICL scores across the three writer/backbone configurations.}
\label{fig:aliasing-fact-count}
\end{figure}
\begin{table*}[t]\centering
\caption{\tablelead{Full-document and target-only writing (Qwen3-4B).} ICL answers with the source text; D2L writes from the full input, while D2L (oracle) writes only from annotated supporting evidence. SQuAD-4 combines four passages and NQ uses full pages.}
\label{tab:mode1}
\small
\setlength{\tabcolsep}{4pt}\renewcommand{\arraystretch}{1.10}
\setlength{\aboverulesep}{0pt}\setlength{\belowrulesep}{0pt}
\begin{tabular}{lrrrrrr}\toprule
\rowcolor{PaperHeader}
& \multicolumn{2}{c}{ICL} & \multicolumn{2}{c}{D2L} & \multicolumn{2}{c}{D2L (oracle)} \\
\rowcolor{PaperHeader}
Task & F1 & R-L & F1 & R-L & F1 & R-L \\\midrule
FactSet-Homo & 1.000 & 1.000 & 0.081 & 0.184 & 0.517 & 0.982 \\
Wikipedia & 0.844 & 0.977 & 0.059 & 0.217 & 0.280 & 0.958 \\
SQuAD-4 & 0.729 & 0.943 & 0.210 & 0.604 & 0.307 & 0.834 \\
NQ & 0.568 & 0.841 & 0.162 & 0.530 & 0.261 & 0.689 \\
NewsQA & 0.479 & 0.770 & 0.217 & 0.538 & 0.351 & 0.787 \\
\bottomrule\end{tabular}
\end{table*}

\paragraph{D2L query-dependent responses.}
Figure~\ref{fig:mode1}b compares D2L-generated and directly optimized LoRA of the same size, with each adapter held fixed across eight questions about its document. The query-dependent activation-energy fraction $\eta$ averages 0.066 and 0.434, respectively. Larger values indicate greater variation in the adapter's direct contribution across queries.

For each of eight synthetic documents $d$, we ask its eight factual queries and append the forced answer prefix ``The $R$ of $S$ is''. Figure~\ref{fig:mode1}b uses only the final token, ``is''. Let $x_{d\ell q}$ denote the input to \texttt{down\_proj} at that position in layer $\ell$, and let $u_{d\ell q}=\Delta W_{d\ell}x_{d\ell q}$ be the document-dependent adapter's direct contribution, including its scaling. We define
\begin{equation}
\begin{aligned}
\bar u_{d\ell}&=\frac{1}{8}\sum_{q=1}^{8}u_{d\ell q},\\
\eta_d&=\frac{\sum_{\ell=1}^{36}\sum_{q=1}^{8}\|u_{d\ell q}-\bar u_{d\ell}\|_2^2}
{\max\!\left(10^{-20},\sum_{\ell=1}^{36}\sum_{q=1}^{8}\|u_{d\ell q}\|_2^2\right)}.
\end{aligned}
\end{equation}
Centering is across queries within each document and layer. The numerator and denominator sum squared coordinates over all 36 layers and eight queries before taking the ratio. The denominator is total uncentered adapter-output energy, not the backbone's activation energy. Reported means are $\eta=\frac{1}{8}\sum_d\eta_d$, giving documents equal weight. The shared D2L bias adapter remains installed during the forward pass but is excluded from $u$. Hidden states therefore reflect the installed model, while the measured vector includes only the direct document-dependent update.

\paragraph{Attention sharing.}
The shared-attention matrix has an off-diagonal fraction of 0.740 to 0.773, versus 7/8 for uniform mixing across eight facts. Across 896 off-diagonal pairs, correlation with competing-answer log-likelihood changes is 0.057.

\paragraph{Content interventions.}
Of 192 Wikipedia birth-year responses, 67 exactly borrow another person's year. Replacing that source's year changes the first output year in 61/67 cases, versus 18/67 when replacing a control source. Averaging its memory values changes the first year in 65/67 cases, versus 22/67 for a control passage. In 5/67 cases, the response follows the replacement year. Borrowed and other passages receive nearly identical mean attention shares (0.119 and 0.117). Privileged target bias raises full-cohort F1/R-L from 0.058/0.219 to 0.255/0.896. This probe uses support annotations.

\subsection{Target Retention: Target Strength and Added Content}
\label{app:dilution-internals}
\paragraph{Synthetic interventions.}
Mean target extraction share falls from 0.503 in isolation to 0.105, 0.057, and 0.035 with 7, 15, and 31 additions. With 31 empty-content sentences, share falls to 0.075 while ROUGE-L Recall remains 0.969, separating attention allocation from harmful content competition.

At fifteen other-type additions, F1/R-L rises from 0.307/0.766 to 0.381/0.945 under amplification, versus 0.248/0.836 under value averaging. Gold-support bias reaches 0.393/0.945 on the same queries. The amplification containment gain is 0.172. Correct and incorrect responses have nearly equal target shares (0.058 and 0.057).

\paragraph{Natural-text validation.}
On the 60-target SQuAD probe with seven distractors, target share falls from 0.873 to 0.117, while enrichment remains near one. With filler, share reaches 0.206 but enrichment rises to 1.471. Both show little correlation with individual answer containment (0.030 and 0.075 at $k=7$).
\begin{figure}[!htbp]\centering
\includegraphics[width=\linewidth]{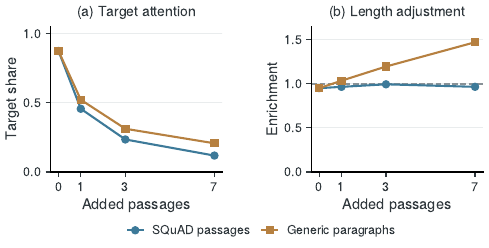}
\caption{Natural-text validation on 60 SQuAD targets. Generic paragraphs omit names, dates, and numbers. Enrichment divides attention share by target token fraction. The dashed line marks 1.}
\end{figure}

\subsection{Response Adaptation: Evidence and Response Control}
\label{app:prompt-internals}
\begin{table}[t]\centering
\caption{\tablelead{Baseline short-answer diagnosis (13 non-LongBench tasks).} The last two columns report the fraction of responses with at most five words, overall and conditional on R-L=1. Conditional rates omit tasks with no full-recall responses.}
\label{tab:prompt-response-control}
\footnotesize
\setlength{\tabcolsep}{2.5pt}\renewcommand{\arraystretch}{1.10}
\setlength{\aboverulesep}{0pt}\setlength{\belowrulesep}{0pt}
\begin{tabular}{>{\raggedright\arraybackslash}p{\dimexpr 0.34000000\linewidth-2\tabcolsep\relax}>{\centering\arraybackslash}p{\dimexpr 0.14000000\linewidth-2\tabcolsep\relax}>{\centering\arraybackslash}p{\dimexpr 0.14000000\linewidth-2\tabcolsep\relax}>{\centering\arraybackslash}p{\dimexpr 0.18000000\linewidth-2\tabcolsep\relax}>{\centering\arraybackslash}p{\dimexpr 0.20000000\linewidth-2\tabcolsep\relax}}\toprule
\rowcolor{PaperHeader}
Method & F1 & R-L & \shortstack{$\leq$5\\words} & \shortstack{Given\\R-L=1} \\\midrule
No-document LM & 0.111 & 0.208 & 0.291 & 0.409 \\
ICL & 0.751 & 0.917 & 0.742 & 0.757 \\
D2L & 0.187 & 0.553 & 0.043 & 0.045 \\
\bottomrule\end{tabular}
\end{table}

\paragraph{Coverage and brevity.}
On FactSet-Homo, 18 responses reach full recall under joint D2L writing, compared with 293 for gold-only D2L and 304 for ICL. Of these, 0, 46, and 304 respectively contain at most five words. Better access to supporting evidence therefore leaves a substantial gap in producing the requested short answer.

\subsection{Capacity Control}
\label{app:frozen-head}
Optimizing D2L's aggregator-output slots with its downstream mapping fixed raises containment from 4/64 to 64/64. Reconstruction from captured latents yields zero recorded relative error.

The analogous SHINE memory-state intervention raises containment from 8/64 to 55/64, versus 57/64 for a directly optimized reference. These supervised probes establish effective writes within the fixed mapping on the tested material.

\subsection{KV Diagnostic Controls}
\begin{figure}[t]\centering
\includegraphics[width=\linewidth]{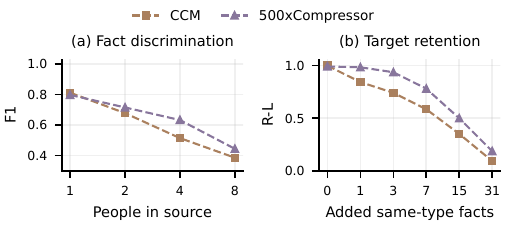}
\caption{Fact discrimination and target retention in existing KV compressors. (a) F1 as a Wikipedia source grows from one to eight people. (b) R-L on 64 fixed targets as same-type facts are added. Both construct 48-position KV without the query.}
\end{figure}
These probes examine fact discrimination and target retention in CCM and 500xCompressor alongside the parameter writer and its no-query control. The CCM and 500xCompressor implementations use the same multi-style teacher data and training schedule as \framework{}. Both construct 48-position KV without the query. Reproduction details are in Appendix~\ref{app:kv-baselines}.

\paragraph{Competing facts.}
\begin{table*}[t]\centering
\caption{\tablelead{Fact discrimination as source count increases.} Birth-year and birthplace queries share a target passage with $K-1$ added people. Pairs report F1 / R-L; borrowed counts cover 152 birth-year queries at $K=8$.}
\small
\setlength{\tabcolsep}{3pt}\renewcommand{\arraystretch}{1.12}
\begin{tabular}{lccccccccc}\toprule
\rowcolor{PaperHeader}
Method & \multicolumn{2}{c}{$K=1$} & \multicolumn{2}{c}{$K=2$} & \multicolumn{2}{c}{$K=4$} & \multicolumn{2}{c}{$K=8$} & Borrowed \\\midrule
500xCompressor & 0.797 & 0.905 & 0.715 & 0.827 & 0.633 & 0.732 & 0.445 & 0.502 & 22/152 \\
CCM & 0.812 & 0.914 & 0.678 & 0.780 & 0.515 & 0.601 & 0.387 & 0.436 & 19/152 \\
\ours{} w/o query & 0.111 & 0.112 & 0.101 & 0.104 & 0.097 & 0.099 & 0.089 & 0.093 & 19/152 \\
\ours{} & 0.825 & 0.931 & 0.815 & 0.927 & 0.825 & 0.934 & 0.819 & 0.934 & 2/152 \\
\bottomrule\end{tabular}
\end{table*}

With one versus eight people in the Wikipedia source, CCM F1 drops from 0.812 to 0.387, and 500xCompressor F1 from 0.797 to 0.445. \ours{} remains stable.

\paragraph{Target retention and evidence selection.}
\begin{table*}[tp]\centering
\caption{\tablelead{Target retention under document expansion.} R-L on 64 targets with $k$ added facts or empty sentences.}
\small
\setlength{\tabcolsep}{3pt}\renewcommand{\arraystretch}{1.12}
\begin{tabular}{llcccccc}\toprule
\rowcolor{PaperHeader}
Method & Additions & $k=0$ & 1 & 3 & 7 & 15 & 31 \\\midrule
\rowcolor{PaperGroup}\multicolumn{8}{l}{500xCompressor} \\
 & Same-type & 0.992 & 0.984 & 0.938 & 0.781 & 0.500 & 0.188 \\
 & Other-type & 0.992 & 1.000 & 0.984 & 0.984 & 0.922 & 0.789 \\
 & Empty & 0.992 & 1.000 & 1.000 & 0.992 & 0.969 & 0.914 \\
\rowcolor{PaperGroup}\multicolumn{8}{l}{CCM} \\
 & Same-type & 1.000 & 0.844 & 0.742 & 0.586 & 0.352 & 0.094 \\
 & Other-type & 1.000 & 0.992 & 0.992 & 0.984 & 0.992 & 0.875 \\
 & Empty & 1.000 & 0.977 & 0.992 & 1.000 & 0.961 & 0.992 \\
\rowcolor{PaperGroup}\multicolumn{8}{l}{\ours{}} \\
 & Same-type & 0.992 & 0.977 & 0.984 & 0.977 & 0.977 & 0.977 \\
 & Other-type & 0.992 & 0.984 & 0.969 & 0.984 & 0.992 & 0.992 \\
 & Empty & 0.992 & 0.977 & 0.977 & 0.984 & 0.992 & 0.992 \\
\bottomrule\end{tabular}
\end{table*}

\begin{table*}[tp]\centering
\caption{\tablelead{Full-document and target-only construction.} Paired F1 scores use the full document / target evidence.}
\label{tab:kv-diagnostic-oracle}
\small
\setlength{\tabcolsep}{3pt}\renewcommand{\arraystretch}{1.12}
\begin{tabular}{lcccccccc}\toprule
\rowcolor{PaperHeader}
Method & \multicolumn{2}{c}{FactSet-Homo} & \multicolumn{2}{c}{Wiki (year)} & \multicolumn{2}{c}{SQuAD-4} & \multicolumn{2}{c}{NQ-4} \\\midrule
500xCompressor & 0.896 & 0.992 & 0.413 & 0.887 & 0.516 & 0.720 & 0.494 & 0.622 \\
CCM & 0.624 & 0.969 & 0.321 & 0.911 & 0.468 & 0.663 & 0.456 & 0.606 \\
\ours{} w/o query & 0.001 & 0.005 & 0.089 & 0.118 & 0.199 & 0.226 & 0.231 & 0.247 \\
\ours{} & 0.970 & 0.972 & 0.928 & 0.968 & 0.711 & 0.759 & 0.609 & 0.657 \\
\bottomrule\end{tabular}
\end{table*}

Same-type additions produce the largest losses for the query-independent learned KV compressors. \ours{} maintains R-L above 0.96 across the tested additions. Target-only construction also improves the query-independent compressors on natural documents, connecting their full-document errors to evidence access. The no-query parameter variant also struggles under target-only construction (Table~\ref{tab:kv-diagnostic-oracle}), so its ablation does not isolate selection among competing facts.

\paragraph{Response adaptation.}
\begin{table*}[tp]\centering
\caption{\tablelead{Response adaptation across memory constructors.} Macro averages over 13 non-LongBench tasks under short-answer and explanation instructions. Length is measured in words.}
\small
\setlength{\tabcolsep}{3pt}\renewcommand{\arraystretch}{1.12}
\begin{tabular}{lcccc}\toprule
\rowcolor{PaperHeader}
Method & \multicolumn{2}{c}{Words} & \multicolumn{2}{c}{R-L} \\\midrule
 & Short & Explain & Short & Explain \\
\midrule
500xCompressor (48) & 4.63 & 46.93 & 0.69 & 0.77 \\
CCM (48) & 4.25 & 46.83 & 0.63 & 0.71 \\
\ours{} & 5.09 & 46.37 & 0.86 & 0.89 \\
\bottomrule\end{tabular}
\end{table*}

The multi-style KV compressors retain the ability to expand their answers under an explanation instruction, including the query-independent models. These results show that the response rigidity observed in D2L does not characterize every memory constructor.

\section{Method Details}
The main writer is defined in Section~\ref{sec:method}; architecture and training settings appear in Appendix~\ref{app:new-method}.

\subsection{Generator and Training Objective}
\label{app:generator-details}
\paragraph{Generator dimensions and scaling.}
For Qwen3-4B, each of the 36 layers supplies eight memory states of width 2,560. Parameter-free LayerNorm precedes a learned layer-specific $2{,}560\!\times\!512$ projection. Each layer then uses one residual MLP block with dimensions $512\!\rightarrow\!2{,}048\!\rightarrow\!512$, SiLU activation, and LayerNorm before and after the two linear transformations. The residual output is normalized to unit L2 norm. A bias-free head at each layer maps the latent vector to 9,728 input-factor and 2,560 output-factor coordinates.

The per-layer, per-rank factor scales are inherited from the released D2L checkpoint and held fixed. After these row scales, the answering implementation applies the checkpoint's multiplier $s=45.254834$ to the factor product, $\Delta W_\ell=s B_\ell A_\ell$, without another division by rank. Section~\ref{sec:method} absorbs this fixed multiplier into the scaled factors. The residual MLP and linear output heads are trained jointly with the encoder. The document-independent D2L bias adapter is omitted.

\paragraph{Exact retention objective.}
Let $p_0$ be the frozen backbone distribution and $p_{\Delta}$ the distribution after applying a generated update. For an unrelated text $u$ of $T_u$ tokens, we use
\begin{equation}
\begin{split}
\ell_{\mathrm{retain}}(u,\Delta)
=\frac{1}{T_u-1}\sum_{t=1}^{T_u-1}
\mathrm{KL}\bigl(&p_0(\cdot\mid u_{\leq t})\,\|\\
&p_{\Delta}(\cdot\mid u_{\leq t})\bigr).
\end{split}
\end{equation}
KL is computed over the full vocabulary at temperature one, averaged over valid next-token positions within each text and then equally over texts. The reference distribution is detached. The student reads the same text without the source document. The training loss adds this average with weight 0.3 to the forward-KL answer-distillation loss, which averages answer positions within each response and then equally across responses.

\paragraph{Retention texts and sampling.}
The fixed training pool contains sixteen unrelated texts: eight PhantomWiki biographies from the 500-person universe and eight LoCoMo dialogue excerpts. Biographies containing names from the 50-person diagnostic universe are excluded. Each source is shuffled with seed 0. Its first twenty texts are reserved for retention evaluation, and the following eight form the training pool. Dialogue excerpts are first limited to 2,000 characters. All retention inputs are tokenized without a chat prompt and limited to their first 128 tokens.

In the default four-worker training run, the sixteen texts are partitioned across workers. Each worker evaluates four texts using the first generated update in its final local microbatch, and worker losses are averaged. Gradients pass through the generated update into the writer. Memory Tokens KV uses the same objective and text pool, replacing the update with its generated memory prefix.

\subsection{Training-Free Variant: \ourstf{}}
\label{app:method}
\ourstf{} is our training-free variant of query-guided writing. It makes an existing writer query-aware without changing its parameters. Given document $D$ and query $q$, a frozen-backbone matcher estimates token relevance $r_i(D,q)$. We smooth and align these scores to document memory positions, then bias the writer's extraction logits.
\begin{equation}
 \boldsymbol{\alpha}_m=\operatorname{softmax}\!\left(\mathbf a_m+\mathbf b(D,q)\right).
 \label{eq:matcher}
\end{equation}
Here $\mathbf a_m$ contains slot $m$'s extraction logits, and $\mathbf b(D,q)$ favors relevant memory positions. The frozen generator converts these states to LoRA.

\paragraph{Bias-only implementation.}
For the smoothed and aligned relevance scores $r_i(D,q)$, the bias in Equation~\ref{eq:matcher} is
\begin{equation}
 b_i(D,q)=\max\!\left\{-c,\tau^{-1}\log\frac{r_i+\epsilon}{\max_j r_j+\epsilon}\right\}.
\end{equation}
The highest-scoring positions receive zero bias, while lower-scoring positions receive negative biases bounded below by $-c$. Temperature $\tau$ controls bias strength; $\epsilon$ stabilizes the logarithm.

The standard Qwen D2L matcher uses 16 attention heads selected on \texttt{calib\_squad\_k4}. Token relevance is smoothed over three neighbors on each side, aligned by character offsets, and converted to log biases with $\tau=1$, $c=12$, and $\epsilon=10^{-6}$. Standard D2L evaluations use reassociated extraction and retain at most 2,048 memory positions by bias before soft aggregation. Unaligned template positions retain their configured neutral bias.

\paragraph{Writer interfaces.}
D2L retains the highest-scoring document positions and adds relevance bias to attention over their projected memories before latent assembly and parameter generation, preserving native scaling and document-independent components. SHINE instead biases memory-token attention to document positions. Its query-dependent memory computation is repeated. Diagnostic comparisons preserve each writer's native scaling and document-independent components. Directly optimized reference adapters use the same parameter shapes. SHINE uses its native prompts with bias strength 4. Transfer tests use matched queries.

Cross-writer results appear in Appendix~\ref{app:current-results}.

\section{Additional Experimental Results}
Settings appear in Appendix~\ref{app:reproducibility}.

\subsection{Comparison with Existing Methods}
\label{app:current-results}
\begin{table*}[!tp]\centering
\caption{\tablelead{Per-task results on Qwen3-4B.} Each method reports F1, R-L, and EM on the same queries. Teal cells mark the best rounded score for each task and metric, excluding ICL.}
\label{tab:current-detailed}
\footnotesize
\setlength{\tabcolsep}{1.5pt}\renewcommand{\arraystretch}{1.16}
\setlength{\aboverulesep}{0pt}\setlength{\belowrulesep}{0pt}
\begin{tabular}{>{\raggedright\arraybackslash}p{\dimexpr.18\textwidth-2\tabcolsep\relax}*{15}{>{\centering\arraybackslash}p{\dimexpr.0546666667\textwidth-2\tabcolsep\relax}}}\toprule
\rowcolor{PaperHeader}
Task & \multicolumn{3}{c}{ICL} & \multicolumn{3}{c}{D2L} & \multicolumn{3}{c}{\shortstack{D2L + Further\\Training}} & \multicolumn{3}{c}{\methodnametf{}} & \multicolumn{3}{c}{\methodname{}} \\
\rowcolor{PaperHeader}
 & F1 & R-L & EM & F1 & R-L & EM & F1 & R-L & EM & F1 & R-L & EM & F1 & R-L & EM \\\midrule
SQuAD & 0.789 & 0.948 & 0.620 & 0.308 & 0.838 & 0.010 & 0.684 & 0.800 & 0.507 & 0.325 & \cellcolor{PaperBest}\textbf{0.881} & 0.013 & \cellcolor{PaperBest}\textbf{0.762} & 0.872 & \cellcolor{PaperBest}\textbf{0.570} \\
DROP & 0.539 & 0.839 & 0.323 & 0.150 & 0.789 & 0.000 & 0.486 & 0.665 & 0.297 & 0.174 & 0.809 & 0.000 & \cellcolor{PaperBest}\textbf{0.602} & \cellcolor{PaperBest}\textbf{0.818} & \cellcolor{PaperBest}\textbf{0.393} \\
Wiki (birth year) & 0.980 & 0.990 & 0.970 & 0.054 & 0.197 & 0.000 & 0.178 & 0.178 & 0.178 & 0.221 & 0.799 & 0.000 & \cellcolor{PaperBest}\textbf{0.928} & \cellcolor{PaperBest}\textbf{0.928} & \cellcolor{PaperBest}\textbf{0.928} \\
Wiki (birthplace) & 0.708 & 0.964 & 0.283 & 0.065 & 0.237 & 0.000 & 0.163 & 0.206 & 0.092 & 0.266 & 0.913 & 0.020 & \cellcolor{PaperBest}\textbf{0.678} & \cellcolor{PaperBest}\textbf{0.921} & \cellcolor{PaperBest}\textbf{0.270} \\
FactSet-Homo & 1.000 & 1.000 & 1.000 & 0.081 & 0.184 & 0.000 & 0.246 & 0.248 & 0.118 & 0.379 & 0.840 & 0.010 & \cellcolor{PaperBest}\textbf{0.970} & \cellcolor{PaperBest}\textbf{0.972} & \cellcolor{PaperBest}\textbf{0.941} \\
FactSet-Hetero & 0.966 & 0.997 & 0.951 & 0.186 & 0.544 & 0.000 & 0.590 & 0.612 & 0.533 & 0.294 & 0.804 & 0.010 & \cellcolor{PaperBest}\textbf{0.893} & \cellcolor{PaperBest}\textbf{0.954} & \cellcolor{PaperBest}\textbf{0.859} \\
FactSet-Distract & 0.978 & 1.000 & 0.967 & 0.317 & 0.776 & 0.118 & 0.745 & 0.762 & 0.720 & 0.395 & 0.952 & 0.138 & \cellcolor{PaperBest}\textbf{0.929} & \cellcolor{PaperBest}\textbf{0.987} & \cellcolor{PaperBest}\textbf{0.895} \\
SQuAD-4 & 0.729 & 0.943 & 0.530 & 0.210 & 0.604 & 0.003 & 0.418 & 0.531 & 0.253 & 0.311 & \cellcolor{PaperBest}\textbf{0.861} & 0.007 & \cellcolor{PaperBest}\textbf{0.711} & \cellcolor{PaperBest}\textbf{0.861} & \cellcolor{PaperBest}\textbf{0.507} \\
SQuAD-Mix & 0.718 & 0.927 & 0.523 & 0.218 & 0.607 & 0.000 & 0.443 & 0.531 & 0.273 & 0.299 & 0.867 & 0.003 & \cellcolor{PaperBest}\textbf{0.744} & \cellcolor{PaperBest}\textbf{0.902} & \cellcolor{PaperBest}\textbf{0.550} \\
NQ-4 & 0.564 & 0.811 & 0.367 & 0.210 & 0.561 & 0.013 & 0.426 & 0.487 & 0.270 & 0.272 & 0.715 & 0.033 & \cellcolor{PaperBest}\textbf{0.609} & \cellcolor{PaperBest}\textbf{0.745} & \cellcolor{PaperBest}\textbf{0.410} \\
NewsQA & 0.479 & 0.770 & 0.243 & 0.217 & 0.538 & 0.007 & 0.342 & 0.466 & 0.183 & 0.273 & \cellcolor{PaperBest}\textbf{0.675} & 0.007 & \cellcolor{PaperBest}\textbf{0.450} & 0.626 & \cellcolor{PaperBest}\textbf{0.257} \\
TriviaQA & 0.748 & 0.892 & 0.657 & 0.249 & 0.785 & 0.077 & 0.652 & 0.783 & 0.560 & 0.287 & 0.855 & 0.093 & \cellcolor{PaperBest}\textbf{0.798} & \cellcolor{PaperBest}\textbf{0.883} & \cellcolor{PaperBest}\textbf{0.717} \\
NQ & 0.568 & 0.841 & 0.393 & 0.162 & 0.530 & 0.007 & 0.380 & 0.478 & 0.233 & 0.218 & 0.664 & 0.013 & \cellcolor{PaperBest}\textbf{0.612} & \cellcolor{PaperBest}\textbf{0.738} & \cellcolor{PaperBest}\textbf{0.433} \\
2Wiki & 0.402 & 0.620 & 0.285 & 0.092 & 0.479 & 0.000 & 0.268 & 0.345 & 0.185 & 0.095 & \cellcolor{PaperBest}\textbf{0.542} & 0.000 & \cellcolor{PaperBest}\textbf{0.419} & 0.512 & \cellcolor{PaperBest}\textbf{0.305} \\
MFQA & 0.495 & 0.734 & 0.187 & 0.264 & 0.451 & 0.020 & 0.255 & 0.362 & 0.027 & 0.371 & \cellcolor{PaperBest}\textbf{0.659} & 0.040 & \cellcolor{PaperBest}\textbf{0.403} & 0.506 & \cellcolor{PaperBest}\textbf{0.107} \\
HotpotQA & 0.550 & 0.671 & 0.405 & 0.114 & 0.423 & 0.025 & 0.292 & 0.370 & 0.185 & 0.162 & \cellcolor{PaperBest}\textbf{0.644} & 0.040 & \cellcolor{PaperBest}\textbf{0.554} & 0.626 & \cellcolor{PaperBest}\textbf{0.400} \\
MuSiQue & 0.189 & 0.426 & 0.110 & 0.052 & 0.197 & 0.000 & 0.082 & 0.140 & 0.015 & 0.085 & \cellcolor{PaperBest}\textbf{0.360} & 0.000 & \cellcolor{PaperBest}\textbf{0.211} & 0.308 & \cellcolor{PaperBest}\textbf{0.130} \\
\bottomrule\end{tabular}
\end{table*}

Table~\ref{tab:current-detailed} reports per-task results for the five methods. On LongBench, \ours{} improves F1 and EM over \ourstf{} on each task while yielding lower reference coverage. Answer accuracy and reference coverage thus differ.

\paragraph{Standalone memory comparison.}
\label{app:memory-token-results}
Table~\ref{tab:memory-token-kv} reports per-task F1, R-L, and EM for \ours{} and the KV baselines in Table~\ref{tab:current-main}. Each uses 6.75\,MiB of document-derived state, represented as a rank-eight LoRA update or 48 KV positions.
\begin{table*}[!tp]\centering
\caption{\tablelead{Standalone memory comparison.} Per-task F1, R-L, and EM for six methods, all using 6.75\,MiB of document-derived state. KV methods use a 48-position budget. Teal cells mark the best rounded score across all six methods for each task and metric.}
\label{tab:memory-token-kv}
\small
\setlength{\tabcolsep}{3pt}\renewcommand{\arraystretch}{1.10}
\setlength{\aboverulesep}{0pt}\setlength{\belowrulesep}{0pt}
\begin{tabular}{>{\raggedright\arraybackslash}p{\dimexpr.22\textwidth-2\tabcolsep\relax}*{9}{>{\centering\arraybackslash}p{\dimexpr.0866666667\textwidth-2\tabcolsep\relax}}}\toprule
\rowcolor{PaperGroup}\multicolumn{10}{l}{\textit{(a) Selected document KV}} \\
\rowcolor{PaperHeader}
Task & \multicolumn{3}{c}{SnapKV} & \multicolumn{3}{c}{DropKV} & \multicolumn{3}{c}{Fast KVzip} \\
\rowcolor{PaperHeader}
 & F1 & R-L & EM & F1 & R-L & EM & F1 & R-L & EM \\\midrule
SQuAD & 0.766 & 0.925 & 0.580 & 0.737 & 0.933 & 0.550 & \cellcolor{PaperBest}\textbf{0.779} & \cellcolor{PaperBest}\textbf{0.942} & \cellcolor{PaperBest}\textbf{0.613} \\
DROP & 0.548 & 0.815 & 0.357 & 0.491 & 0.797 & 0.297 & 0.529 & 0.756 & 0.340 \\
Wiki (year) & 0.975 & 0.980 & 0.970 & \cellcolor{PaperBest}\textbf{0.986} & \cellcolor{PaperBest}\textbf{0.990} & \cellcolor{PaperBest}\textbf{0.984} & 0.711 & 0.737 & 0.704 \\
Wiki (place) & \cellcolor{PaperBest}\textbf{0.692} & 0.947 & 0.260 & 0.668 & \cellcolor{PaperBest}\textbf{0.951} & 0.211 & 0.167 & 0.345 & 0.076 \\
FactSet-Homo & 0.972 & 0.972 & 0.947 & 0.997 & 0.997 & \cellcolor{PaperBest}\textbf{0.997} & \cellcolor{PaperBest}\textbf{0.998} & \cellcolor{PaperBest}\textbf{1.000} & \cellcolor{PaperBest}\textbf{0.997} \\
FactSet-Hetero & 0.947 & 0.975 & 0.928 & 0.940 & \cellcolor{PaperBest}\textbf{0.997} & 0.911 & \cellcolor{PaperBest}\textbf{0.953} & \cellcolor{PaperBest}\textbf{0.997} & \cellcolor{PaperBest}\textbf{0.931} \\
FactSet-Distract & \cellcolor{PaperBest}\textbf{0.959} & 0.992 & \cellcolor{PaperBest}\textbf{0.934} & 0.935 & \cellcolor{PaperBest}\textbf{1.000} & 0.905 & 0.927 & \cellcolor{PaperBest}\textbf{1.000} & 0.891 \\
SQuAD-4 & \cellcolor{PaperBest}\textbf{0.713} & \cellcolor{PaperBest}\textbf{0.931} & \cellcolor{PaperBest}\textbf{0.517} & 0.670 & 0.903 & 0.467 & 0.257 & 0.439 & 0.143 \\
SQuAD-Mix & 0.688 & 0.898 & 0.483 & 0.645 & 0.867 & 0.463 & 0.259 & 0.479 & 0.123 \\
NQ-4 & 0.507 & \cellcolor{PaperBest}\textbf{0.758} & 0.290 & 0.484 & 0.738 & 0.290 & 0.218 & 0.395 & 0.107 \\
NewsQA & 0.461 & 0.701 & 0.217 & \cellcolor{PaperBest}\textbf{0.470} & \cellcolor{PaperBest}\textbf{0.705} & 0.223 & 0.075 & 0.176 & 0.030 \\
TriviaQA & 0.754 & \cellcolor{PaperBest}\textbf{0.891} & 0.660 & 0.763 & 0.877 & 0.677 & 0.467 & 0.591 & 0.397 \\
NQ & 0.485 & \cellcolor{PaperBest}\textbf{0.743} & 0.323 & 0.497 & 0.711 & 0.320 & 0.156 & 0.309 & 0.067 \\
2Wiki & 0.302 & 0.412 & 0.220 & 0.305 & 0.421 & 0.200 & 0.232 & 0.328 & 0.165 \\
MFQA & 0.407 & 0.599 & 0.127 & \cellcolor{PaperBest}\textbf{0.420} & \cellcolor{PaperBest}\textbf{0.613} & \cellcolor{PaperBest}\textbf{0.153} & 0.226 & 0.333 & 0.027 \\
HotpotQA & 0.484 & 0.583 & 0.340 & 0.401 & 0.515 & 0.290 & 0.250 & 0.367 & 0.155 \\
MuSiQue & 0.152 & 0.299 & 0.090 & 0.142 & 0.241 & 0.085 & 0.101 & 0.208 & 0.040 \\
\midrule
Macro & 0.636 & \cellcolor{PaperBest}\textbf{0.789} & 0.485 & 0.621 & 0.780 & 0.472 & 0.430 & 0.553 & 0.341 \\
\midrule\addlinespace[5pt]
\rowcolor{PaperGroup}\multicolumn{10}{l}{\textit{(b) Learned KV and LoRA}} \\
\rowcolor{PaperHeader}
Task & \multicolumn{3}{c}{500xCompressor} & \multicolumn{3}{c}{CCM} & \multicolumn{3}{c}{\methodname{}} \\
\rowcolor{PaperHeader}
 & F1 & R-L & EM & F1 & R-L & EM & F1 & R-L & EM \\\midrule
SQuAD & 0.719 & 0.826 & 0.530 & 0.672 & 0.771 & 0.483 & 0.762 & 0.872 & 0.570 \\
DROP & 0.523 & 0.695 & 0.337 & 0.517 & 0.680 & 0.337 & \cellcolor{PaperBest}\textbf{0.602} & \cellcolor{PaperBest}\textbf{0.818} & \cellcolor{PaperBest}\textbf{0.393} \\
Wiki (year) & 0.413 & 0.414 & 0.411 & 0.321 & 0.322 & 0.319 & 0.928 & 0.928 & 0.928 \\
Wiki (place) & 0.449 & 0.593 & 0.197 & 0.364 & 0.471 & 0.148 & 0.678 & 0.921 & \cellcolor{PaperBest}\textbf{0.270} \\
FactSet-Homo & 0.896 & 0.896 & 0.875 & 0.624 & 0.630 & 0.566 & 0.970 & 0.972 & 0.941 \\
FactSet-Hetero & 0.889 & 0.956 & 0.849 & 0.848 & 0.893 & 0.806 & 0.893 & 0.954 & 0.859 \\
FactSet-Distract & 0.830 & 0.942 & 0.776 & 0.892 & 0.951 & 0.852 & 0.929 & 0.987 & 0.895 \\
SQuAD-4 & 0.516 & 0.630 & 0.333 & 0.468 & 0.558 & 0.280 & 0.711 & 0.861 & 0.507 \\
SQuAD-Mix & 0.515 & 0.629 & 0.337 & 0.526 & 0.612 & 0.350 & \cellcolor{PaperBest}\textbf{0.744} & \cellcolor{PaperBest}\textbf{0.902} & \cellcolor{PaperBest}\textbf{0.550} \\
NQ-4 & 0.494 & 0.583 & 0.333 & 0.456 & 0.549 & 0.290 & \cellcolor{PaperBest}\textbf{0.609} & 0.745 & \cellcolor{PaperBest}\textbf{0.410} \\
NewsQA & 0.388 & 0.509 & 0.207 & 0.401 & 0.507 & 0.203 & 0.450 & 0.626 & \cellcolor{PaperBest}\textbf{0.257} \\
TriviaQA & 0.726 & 0.796 & 0.650 & 0.716 & 0.791 & 0.647 & \cellcolor{PaperBest}\textbf{0.798} & 0.883 & \cellcolor{PaperBest}\textbf{0.717} \\
NQ & 0.452 & 0.532 & 0.293 & 0.424 & 0.505 & 0.260 & \cellcolor{PaperBest}\textbf{0.612} & 0.738 & \cellcolor{PaperBest}\textbf{0.433} \\
2Wiki & 0.250 & 0.315 & 0.175 & 0.282 & 0.340 & 0.205 & \cellcolor{PaperBest}\textbf{0.419} & \cellcolor{PaperBest}\textbf{0.512} & \cellcolor{PaperBest}\textbf{0.305} \\
MFQA & 0.288 & 0.385 & 0.040 & 0.274 & 0.375 & 0.047 & 0.403 & 0.506 & 0.107 \\
HotpotQA & 0.342 & 0.395 & 0.235 & 0.319 & 0.359 & 0.230 & \cellcolor{PaperBest}\textbf{0.554} & \cellcolor{PaperBest}\textbf{0.626} & \cellcolor{PaperBest}\textbf{0.400} \\
MuSiQue & 0.156 & 0.200 & 0.080 & 0.136 & 0.193 & 0.065 & \cellcolor{PaperBest}\textbf{0.211} & \cellcolor{PaperBest}\textbf{0.308} & \cellcolor{PaperBest}\textbf{0.130} \\
\midrule
Macro & 0.520 & 0.606 & 0.392 & 0.485 & 0.559 & 0.358 & \cellcolor{PaperBest}\textbf{0.663} & 0.774 & \cellcolor{PaperBest}\textbf{0.510} \\
\bottomrule\end{tabular}
\end{table*}

\paragraph{Evaluation across writers.}
\begin{table}[t]\centering
\caption{\tablelead{\ourstf{} across writers.} Macro F1 over 13 non-LongBench tasks: three FactSet variants, two Wikipedia attributes, SQuAD, DROP, NewsQA, TriviaQA, NQ, SQuAD-4, SQuAD-Mix, and NQ-4. Tasks have equal weight; queries are matched per writer.}
\label{tab:cross-writer}
\footnotesize
\setlength{\tabcolsep}{2.5pt}\renewcommand{\arraystretch}{1.10}
\setlength{\aboverulesep}{0pt}\setlength{\belowrulesep}{0pt}
\begin{tabular}{>{\raggedright\arraybackslash}p{\dimexpr 0.45000000\linewidth-2\tabcolsep\relax}>{\centering\arraybackslash}p{\dimexpr 0.22000000\linewidth-2\tabcolsep\relax}>{\centering\arraybackslash}p{\dimexpr 0.33000000\linewidth-2\tabcolsep\relax}}\toprule
\rowcolor{PaperHeader}
Writer / backbone & raw & \methodnametf{} \\
\midrule
D2L / Qwen3-4B & 0.19 & \cellcolor{PaperBest}\textbf{0.29} \\
D2L / Gemma-2-2B & 0.51 & \cellcolor{PaperBest}\textbf{0.73} \\
SHINE / Qwen3-8B & 0.51 & \cellcolor{PaperBest}\textbf{0.64} \\
\bottomrule\end{tabular}
\end{table}

Table~\ref{tab:cross-writer} summarizes \ourstf{} across writers on the common thirteen-task cohort. Tasks receive equal weight, and queries are matched within each writer. Table~\ref{tab:current-main} reports the full 17-task Qwen average.

\paragraph{Gemma Replication}
\begin{table*}[!tp]\centering
\caption{\tablelead{Per-task benchmark results on Gemma-2-2B.} \ours{} uses the Gemma teacher with the same objective and training budget as the Qwen configuration.}
\label{tab:gemma-detailed}
\footnotesize
\setlength{\tabcolsep}{3pt}\renewcommand{\arraystretch}{1.10}
\setlength{\aboverulesep}{0pt}\setlength{\belowrulesep}{0pt}
\begin{tabular}{>{\raggedright\arraybackslash}p{\dimexpr 0.22000000\textwidth-2\tabcolsep\relax}>{\centering\arraybackslash}p{\dimexpr 0.06500000\textwidth-2\tabcolsep\relax}>{\centering\arraybackslash}p{\dimexpr 0.06500000\textwidth-2\tabcolsep\relax}>{\centering\arraybackslash}p{\dimexpr 0.06500000\textwidth-2\tabcolsep\relax}>{\centering\arraybackslash}p{\dimexpr 0.06500000\textwidth-2\tabcolsep\relax}>{\centering\arraybackslash}p{\dimexpr 0.06500000\textwidth-2\tabcolsep\relax}>{\centering\arraybackslash}p{\dimexpr 0.06500000\textwidth-2\tabcolsep\relax}>{\centering\arraybackslash}p{\dimexpr 0.06500000\textwidth-2\tabcolsep\relax}>{\centering\arraybackslash}p{\dimexpr 0.06500000\textwidth-2\tabcolsep\relax}>{\centering\arraybackslash}p{\dimexpr 0.06500000\textwidth-2\tabcolsep\relax}>{\centering\arraybackslash}p{\dimexpr 0.06500000\textwidth-2\tabcolsep\relax}>{\centering\arraybackslash}p{\dimexpr 0.06500000\textwidth-2\tabcolsep\relax}>{\centering\arraybackslash}p{\dimexpr 0.06500000\textwidth-2\tabcolsep\relax}}\toprule
\rowcolor{PaperHeader}
& \multicolumn{3}{c}{\textbf{ICL}} & \multicolumn{3}{c}{\textbf{D2L}} & \multicolumn{3}{c}{\methodnametf{}} & \multicolumn{3}{c}{\methodname{}} \\
\rowcolor{PaperHeader}
Task & F1 & R-L & EM & F1 & R-L & EM & F1 & R-L & EM & F1 & R-L & EM \\\midrule
SQuAD & 0.882 & 0.917 & 0.753 & 0.769 & 0.799 & 0.633 & 0.828 & 0.848 & \cellcolor{PaperBest}\textbf{0.700} & \cellcolor{PaperBest}\textbf{0.841} & \cellcolor{PaperBest}\textbf{0.861} & 0.690 \\
DROP & 0.691 & 0.754 & 0.553 & 0.535 & 0.587 & 0.433 & 0.601 & 0.639 & 0.483 & \cellcolor{PaperBest}\textbf{0.686} & \cellcolor{PaperBest}\textbf{0.747} & \cellcolor{PaperBest}\textbf{0.530} \\
Wiki (birth year) & 0.969 & 0.970 & 0.967 & 0.181 & 0.181 & 0.181 & 0.559 & 0.559 & 0.559 & \cellcolor{PaperBest}\textbf{0.931} & \cellcolor{PaperBest}\textbf{0.931} & \cellcolor{PaperBest}\textbf{0.931} \\
Wiki (birthplace) & 0.712 & 0.964 & 0.283 & 0.174 & 0.237 & 0.059 & 0.624 & 0.839 & 0.243 & \cellcolor{PaperBest}\textbf{0.713} & \cellcolor{PaperBest}\textbf{0.936} & \cellcolor{PaperBest}\textbf{0.303} \\
FactSet-Homo & 1.000 & 1.000 & 1.000 & 0.418 & 0.426 & 0.319 & 0.938 & 0.939 & 0.891 & \cellcolor{PaperBest}\textbf{0.996} & \cellcolor{PaperBest}\textbf{0.997} & \cellcolor{PaperBest}\textbf{0.990} \\
FactSet-Hetero & 1.000 & 1.000 & 1.000 & 0.696 & 0.701 & 0.648 & \cellcolor{PaperBest}\textbf{0.947} & 0.952 & \cellcolor{PaperBest}\textbf{0.938} & 0.934 & \cellcolor{PaperBest}\textbf{0.961} & 0.931 \\
FactSet-Distract & 1.000 & 1.000 & 1.000 & 0.847 & 0.847 & 0.826 & \cellcolor{PaperBest}\textbf{0.989} & \cellcolor{PaperBest}\textbf{0.990} & \cellcolor{PaperBest}\textbf{0.984} & 0.972 & 0.979 & 0.970 \\
SQuAD-4 & 0.873 & 0.910 & 0.753 & 0.517 & 0.551 & 0.377 & 0.793 & \cellcolor{PaperBest}\textbf{0.840} & 0.647 & \cellcolor{PaperBest}\textbf{0.817} & 0.838 & \cellcolor{PaperBest}\textbf{0.670} \\
SQuAD-Mix & 0.868 & 0.904 & 0.713 & 0.514 & 0.558 & 0.333 & 0.788 & 0.821 & 0.633 & \cellcolor{PaperBest}\textbf{0.824} & \cellcolor{PaperBest}\textbf{0.853} & \cellcolor{PaperBest}\textbf{0.657} \\
NQ-4 & 0.701 & 0.726 & 0.507 & 0.476 & 0.485 & 0.310 & 0.589 & 0.601 & 0.420 & \cellcolor{PaperBest}\textbf{0.674} & \cellcolor{PaperBest}\textbf{0.680} & \cellcolor{PaperBest}\textbf{0.470} \\
NewsQA & 0.606 & 0.642 & 0.387 & 0.415 & 0.456 & 0.233 & 0.537 & 0.562 & 0.337 & \cellcolor{PaperBest}\textbf{0.561} & \cellcolor{PaperBest}\textbf{0.573} & \cellcolor{PaperBest}\textbf{0.370} \\
TriviaQA & 0.820 & 0.886 & 0.730 & 0.696 & 0.746 & 0.623 & 0.762 & 0.808 & 0.693 & \cellcolor{PaperBest}\textbf{0.828} & \cellcolor{PaperBest}\textbf{0.883} & \cellcolor{PaperBest}\textbf{0.747} \\
NQ & 0.683 & 0.712 & 0.517 & 0.437 & 0.459 & 0.270 & 0.501 & 0.523 & 0.350 & \cellcolor{PaperBest}\textbf{0.636} & \cellcolor{PaperBest}\textbf{0.645} & \cellcolor{PaperBest}\textbf{0.467} \\
\midrule
\rowcolor{PaperGroup}
\textbf{Average} & 0.831 & 0.876 & 0.705 & 0.513 & 0.541 & 0.404 & 0.727 & 0.763 & 0.606 & \cellcolor{PaperBest}\textbf{0.801} & \cellcolor{PaperBest}\textbf{0.837} & \cellcolor{PaperBest}\textbf{0.671} \\
\bottomrule\end{tabular}
\end{table*}

Table~\ref{tab:gemma-detailed} reports per-task results for the separately trained Gemma model.

\paragraph{Comparison with OSCAR on Mistral}
\label{app:mistral-oscar}
The main-text Table~\ref{tab:mistral-oscar} compares \ours{} on Mistral with released OSCAR \citep{louis2025oscar}. \ours{} has higher F1 and EM with less average document-derived state. OSCAR has similar reference coverage and produces longer answers, so the difference in exact and token-level correctness should be read alongside answer length. We use the official checkpoint (Appendix~\ref{app:oscar-protocol}).

\subsection{Memory Recall under Intervening Context}
Figure~\ref{fig:intervening-text} tests whether fixed query-specific document memory remains usable after increasing amounts of unrelated text. Here, memory recall refers to using the constructed memory to answer the original question, rather than locating evidence in the full source document. Tasks, prompts, and insertion procedures are specified in Appendix~\ref{app:intervening-protocol}.

\paragraph{Accuracy and answer expression.}
Figure~\ref{fig:intervening-text} reports macro F1 and EM. From zero insertion to 32K, \ours{} macro F1 changes from 0.735 to 0.611, EM from 0.599 to 0.478, and R-L from 0.839 to 0.771. ICL changes from 0.745 to 0.610, 0.606 to 0.422, and 0.907 to 0.874, respectively. DropKV changes from 0.719 to 0.554 in F1, 0.583 to 0.375 in EM, and 0.884 to 0.808 in R-L. Memory Tokens KV falls from 0.773 to 0.050 in F1, from 0.635 to 0.016 in EM, and from 0.899 to 0.144 in R-L. Its loss therefore includes a substantial decline in reference coverage, alongside an increase in answer length. Without LoRA or the document, macro F1 is only 0.037 at 32K.

The response changes differ across tasks. On FactSet-Homo, \ours{} mean answer length changes from 2.0 to 2.1 words, while ICL increases from 2.0 to 5.7. ICL R-L remains 0.990 at 32K, but EM falls to 0.454. Among its 166 newly incorrect exact-match responses, 163 retain full R-L coverage. DropKV shows a similar pattern: mean answer length rises from 2.0 to 5.6 words, while R-L remains 0.985 and EM falls to 0.385 at 32K. These patterns are consistent with extra answer text contributing to the exact-match loss. Thus, the aggregate results include both task-dependent accuracy changes and changes in answer expression. The query is known during construction; we test later use of that fixed update.

\begin{table*}[t]\centering
\caption{\tablelead{Position control under intervening context.} Each cell reports F1 / R-L. RoPE-aligned KV shifts only the learned memory keys' RoPE positions to restore their relative distance to the question without removing intervening text. Original KV and \ours{} scores are from Figure~\ref{fig:intervening-text}; RoPE-aligned KV scores are from the additional position control.}
\label{tab:intervening-position}
\footnotesize
\setlength{\tabcolsep}{4pt}\renewcommand{\arraystretch}{1.10}
\begin{tabular}{@{}lcccccc@{}}\toprule
\rowcolor{PaperHeader}
 & \multicolumn{3}{c}{4K intervening tokens} & \multicolumn{3}{c}{32K intervening tokens} \\
\rowcolor{PaperHeader}
Task & \shortstack{Memory Tokens\\KV} & \shortstack{RoPE-aligned\\KV} & \ours{} & \shortstack{Memory Tokens\\KV} & \shortstack{RoPE-aligned\\KV} & \ours{} \\\midrule
SQuAD-4 & .316 / .502 & .647 / .859 & .591 / .812 & .054 / .201 & .087 / .290 & .574 / .811 \\
NQ-4 & .256 / .429 & .446 / .625 & .456 / .659 & .115 / .295 & .116 / .316 & .405 / .648 \\
Wikipedia-Year & .243 / .260 & .893 / .898 & .910 / .918 & .012 / .026 & .039 / .125 & .875 / .875 \\
FactSet-Homo & .269 / .280 & .745 / .788 & .957 / .957 & .002 / .013 & .004 / .018 & .947 / .947 \\
NewsQA & .192 / .308 & .312 / .626 & .349 / .565 & .028 / .088 & .056 / .150 & .316 / .529 \\
SQuAD-Mix & .336 / .514 & .638 / .845 & .620 / .838 & .092 / .240 & .109 / .325 & .547 / .817 \\
\bottomrule\end{tabular}
\end{table*}

\paragraph{Restoring positional proximity helps at 4K but is insufficient at 32K.}
The RoPE alignment control restores the memory-to-question RoPE offsets of the no-insertion condition while retaining the full intervening sequence (Appendix~\ref{app:position-control}). At 4K, it substantially improves Memory Tokens KV on all six tasks and exceeds LoRA in F1 on SQuAD-4 and SQuAD-Mix. At 32K, its F1 remains between 0.004 and 0.116, well below LoRA on every task (Table~\ref{tab:intervening-position}). Thus, correcting this positional mismatch addresses an important source of degradation at 4K but does not explain away the long-interference gap. Competition with intervening tokens for attention remains a plausible explanation, rather than a directly measured mechanism.

\paragraph{Memory access and parameter updates.}
Memory Tokens KV requires attention to a fixed memory prefix, so additional context may interfere with access to the stored information even though the memory itself is unchanged. Generated LoRA modifies feed-forward computation at each subsequent position without requiring memory-prefix retrieval, which may support more stable recall.

Two findings support the usefulness of this parameter pathway. First, at 32K insertion, LoRA retains F1 of 0.611 and R-L of 0.771, compared with 0.050 and 0.144 for Memory Tokens KV. Second, with identical prompts on the first 200 GSM8K questions, loading LoRA generated from unrelated documents yields 92\% accuracy versus 94\% without LoRA (Appendix~\ref{app:retention-behavior}). The first result measures access to the constructed document memory. The second measures preservation of unrelated reasoning performance after loading LoRA.

\subsection{Diagnostic Improvements}

\label{app:targeted-repair}
We revisit the fact-discrimination and target-retention probes from Section~\ref{sec:diagnosis} to test repair through improved evidence access, comparing D2L, \ourstf{}, and \ours{}.
\paragraph{Borrowed-answer correction.}
On the 304-query Wikipedia repair cohort, D2L yields 60 correct, 119 borrowed, and 125 other responses. \ourstf{} yields 243, 30, and 31. Of 119 borrowed cases, 92 become correct, 19 stay borrowed, and eight become other errors.

\ours{} yields 282 correct, three borrowed, and 19 other responses. The corresponding counts are 301/1/2 for ICL and 259/4/41 for the original-D2L-teacher control. Of the same 119 baseline-borrowed cases, 113 become correct and six become other errors. Its Token F1 gain over D2L is 0.874, and over \ourstf{} is 0.706. This measures correct recovery as well as removal of borrowed answers on the repair cohort.

\paragraph{Fixed-target expansion.}
\begin{table*}[!tp]\centering
\caption{\tablelead{Diagnostic repair under fixed-target expansion.} All conditions retain the same 64 target queries and references. $k$ counts added same-type facts, other-type facts, or empty units. The target-only result at $k=0$ is shared across content types. Bold, shaded values mark the best method for each content type, metric, and $k$.}
\label{tab:current-dilution-repair}
\small
\setlength{\tabcolsep}{3pt}\renewcommand{\arraystretch}{1.10}
\setlength{\aboverulesep}{0pt}\setlength{\belowrulesep}{0pt}
\begin{tabular}{>{\raggedright\arraybackslash}p{\dimexpr 0.24000000\textwidth-2\tabcolsep\relax}>{\centering\arraybackslash}p{\dimexpr 0.10000000\textwidth-2\tabcolsep\relax}>{\centering\arraybackslash}p{\dimexpr 0.11000000\textwidth-2\tabcolsep\relax}>{\centering\arraybackslash}p{\dimexpr 0.11000000\textwidth-2\tabcolsep\relax}>{\centering\arraybackslash}p{\dimexpr 0.11000000\textwidth-2\tabcolsep\relax}>{\centering\arraybackslash}p{\dimexpr 0.11000000\textwidth-2\tabcolsep\relax}>{\centering\arraybackslash}p{\dimexpr 0.11000000\textwidth-2\tabcolsep\relax}>{\centering\arraybackslash}p{\dimexpr 0.11000000\textwidth-2\tabcolsep\relax}}\toprule
\rowcolor{PaperHeader}
Added content & Metric & $k=0$ & $k=1$ & $k=3$ & $k=7$ & $k=15$ & $k=31$ \\\midrule
\rowcolor{PaperGroup}\multicolumn{8}{l}{\textit{D2L}} \\
Same-type facts & F1 & 0.407 & 0.303 & 0.188 & 0.036 & 0.020 & 0.004 \\
 & R-L & 0.984 & 0.742 & 0.469 & 0.125 & 0.062 & 0.016 \\
 & EM & 0.141 & 0.094 & 0.062 & 0.000 & 0.000 & 0.000 \\
Other-type facts & F1 & 0.407 & 0.406 & 0.374 & 0.346 & 0.311 & 0.304 \\
 & R-L & 0.984 & 0.977 & 0.914 & 0.844 & 0.781 & 0.758 \\
 & EM & 0.141 & 0.141 & 0.125 & 0.125 & 0.109 & 0.109 \\
Empty content & F1 & 0.407 & 0.410 & 0.410 & 0.415 & 0.386 & 0.378 \\
 & R-L & 0.984 & \cellcolor{PaperBest}\textbf{0.984} & \cellcolor{PaperBest}\textbf{0.984} & \cellcolor{PaperBest}\textbf{1.000} & 0.961 & 0.969 \\
 & EM & 0.141 & 0.141 & 0.141 & 0.141 & 0.125 & 0.109 \\
\rowcolor{PaperGroup}\multicolumn{8}{l}{\textit{\methodnametf{}}} \\
Same-type facts & F1 & 0.395 & 0.352 & 0.354 & 0.322 & 0.294 & 0.198 \\
 & R-L & 0.953 & 0.781 & 0.836 & 0.797 & 0.727 & 0.617 \\
 & EM & 0.141 & 0.156 & 0.141 & 0.109 & 0.094 & 0.016 \\
Other-type facts & F1 & 0.395 & 0.396 & 0.385 & 0.400 & 0.381 & 0.366 \\
 & R-L & 0.953 & 0.961 & 0.945 & 0.961 & 0.945 & 0.938 \\
 & EM & 0.141 & 0.141 & 0.125 & 0.141 & 0.125 & 0.109 \\
Empty content & F1 & 0.395 & 0.396 & 0.379 & 0.398 & 0.376 & 0.379 \\
 & R-L & 0.953 & 0.961 & 0.945 & 0.953 & 0.938 & 0.945 \\
 & EM & 0.141 & 0.141 & 0.125 & 0.141 & 0.125 & 0.125 \\
\rowcolor{PaperGroup}\multicolumn{8}{l}{\textit{\methodname{}}} \\
Same-type facts & F1 & \cellcolor{PaperBest}\textbf{0.955} & \cellcolor{PaperBest}\textbf{0.935} & \cellcolor{PaperBest}\textbf{0.955} & \cellcolor{PaperBest}\textbf{0.958} & \cellcolor{PaperBest}\textbf{0.947} & \cellcolor{PaperBest}\textbf{0.948} \\
 & R-L & \cellcolor{PaperBest}\textbf{0.992} & \cellcolor{PaperBest}\textbf{0.977} & \cellcolor{PaperBest}\textbf{0.984} & \cellcolor{PaperBest}\textbf{0.977} & \cellcolor{PaperBest}\textbf{0.977} & \cellcolor{PaperBest}\textbf{0.977} \\
 & EM & \cellcolor{PaperBest}\textbf{0.922} & \cellcolor{PaperBest}\textbf{0.906} & \cellcolor{PaperBest}\textbf{0.922} & \cellcolor{PaperBest}\textbf{0.938} & \cellcolor{PaperBest}\textbf{0.922} & \cellcolor{PaperBest}\textbf{0.922} \\
Other-type facts & F1 & \cellcolor{PaperBest}\textbf{0.955} & \cellcolor{PaperBest}\textbf{0.937} & \cellcolor{PaperBest}\textbf{0.952} & \cellcolor{PaperBest}\textbf{0.958} & \cellcolor{PaperBest}\textbf{0.955} & \cellcolor{PaperBest}\textbf{0.971} \\
 & R-L & \cellcolor{PaperBest}\textbf{0.992} & \cellcolor{PaperBest}\textbf{0.984} & \cellcolor{PaperBest}\textbf{0.969} & \cellcolor{PaperBest}\textbf{0.984} & \cellcolor{PaperBest}\textbf{0.992} & \cellcolor{PaperBest}\textbf{0.992} \\
 & EM & \cellcolor{PaperBest}\textbf{0.922} & \cellcolor{PaperBest}\textbf{0.906} & \cellcolor{PaperBest}\textbf{0.938} & \cellcolor{PaperBest}\textbf{0.938} & \cellcolor{PaperBest}\textbf{0.938} & \cellcolor{PaperBest}\textbf{0.953} \\
Empty content & F1 & \cellcolor{PaperBest}\textbf{0.955} & \cellcolor{PaperBest}\textbf{0.925} & \cellcolor{PaperBest}\textbf{0.893} & \cellcolor{PaperBest}\textbf{0.909} & \cellcolor{PaperBest}\textbf{0.864} & \cellcolor{PaperBest}\textbf{0.874} \\
 & R-L & \cellcolor{PaperBest}\textbf{0.992} & 0.977 & 0.977 & 0.984 & \cellcolor{PaperBest}\textbf{0.992} & \cellcolor{PaperBest}\textbf{0.992} \\
 & EM & \cellcolor{PaperBest}\textbf{0.922} & \cellcolor{PaperBest}\textbf{0.891} & \cellcolor{PaperBest}\textbf{0.844} & \cellcolor{PaperBest}\textbf{0.875} & \cellcolor{PaperBest}\textbf{0.797} & \cellcolor{PaperBest}\textbf{0.812} \\
\bottomrule\end{tabular}
\end{table*}

Table~\ref{tab:current-dilution-repair} fixes 64 target queries and references while adding same-type facts, other-type facts, or empty content. It directly tests the target-retention claim in Section~\ref{sec:experiments}. With 31 same-type additions, F1 falls by 0.403 for D2L and 0.197 with matching, versus 0.007 for \ours{}. \ours{} also maintains F1 and EM under other-type fact additions. With empty-content additions, it preserves reference coverage, while F1 and EM decrease. Inspection of paired outputs at $k=31$ shows that all eight newly failing exact-match responses expand correct short answers into full sentences.

\paragraph{Relation to response adaptation.}
The paired aliasing and dilution probes test recovery of target answers under controlled changes to source content. The prompt controls in Appendix~\ref{app:prompt-internals} test a separate requirement: whether the answerer can use this evidence under different response instructions. The empty-content results further distinguish target-answer preservation from compliance with the requested answer form.

\paragraph{Response Adaptation Results}
\paragraph{Matched instructions.}
Across the 17 tasks, \ours{} produces means of 7.2, 49.3, 13.2, and 26.0 words under short-answer, explanation, sentence, and query-only prompts, respectively.

\paragraph{Matched response instructions and writer scope.}
\begin{table*}[!tp]\centering
\caption{\tablelead{Answer quality and response length under four instructions.} All methods use the same 13 tasks and three training styles plus a query-only instruction. Length is measured in words.}
\label{tab:prompt-seen}
\small
\setlength{\tabcolsep}{3pt}\renewcommand{\arraystretch}{1.10}
\setlength{\aboverulesep}{0pt}\setlength{\belowrulesep}{0pt}
\begin{tabular}{>{\raggedright\arraybackslash}p{\dimexpr 0.28000000\textwidth-2\tabcolsep\relax}>{\centering\arraybackslash}p{\dimexpr 0.18000000\textwidth-2\tabcolsep\relax}>{\centering\arraybackslash}p{\dimexpr 0.18000000\textwidth-2\tabcolsep\relax}>{\centering\arraybackslash}p{\dimexpr 0.18000000\textwidth-2\tabcolsep\relax}>{\centering\arraybackslash}p{\dimexpr 0.18000000\textwidth-2\tabcolsep\relax}}\toprule
\rowcolor{PaperHeader}
Method & F1 & R-L & EM & Mean words \\\midrule
\rowcolor{PaperGroup}\multicolumn{5}{l}{\textit{Short answer}} \\
ICL & 0.751 & 0.917 & 0.602 & 7.9 \\
D2L & 0.187 & 0.553 & 0.018 & 15.1 \\
\methodname{} & \cellcolor{PaperBest}\textbf{0.745} & \cellcolor{PaperBest}\textbf{0.862} & \cellcolor{PaperBest}\textbf{0.594} & 5.1 \\
\addlinespace[3pt]
\rowcolor{PaperGroup}\multicolumn{5}{l}{\textit{One sentence}} \\
ICL & 0.306 & 0.927 & 0.000 & 13.1 \\
D2L & 0.177 & 0.550 & 0.000 & 12.4 \\
\methodname{} & \cellcolor{PaperBest}\textbf{0.297} & \cellcolor{PaperBest}\textbf{0.874} & \cellcolor{PaperBest}\textbf{0.001} & 11.3 \\
\addlinespace[3pt]
\rowcolor{PaperGroup}\multicolumn{5}{l}{\textit{Query only}} \\
ICL & 0.187 & 0.959 & 0.000 & 53.9 \\
D2L & 0.146 & 0.577 & 0.000 & 26.8 \\
\methodname{} & \cellcolor{PaperBest}\textbf{0.275} & \cellcolor{PaperBest}\textbf{0.885} & \cellcolor{PaperBest}\textbf{0.004} & 20.1 \\
\addlinespace[3pt]
\rowcolor{PaperGroup}\multicolumn{5}{l}{\textit{Explanation}} \\
ICL & 0.092 & 0.955 & 0.000 & 49.9 \\
D2L & \cellcolor{PaperBest}\textbf{0.131} & 0.578 & \cellcolor{PaperBest}\textbf{0.000} & 25.4 \\
\methodname{} & 0.086 & \cellcolor{PaperBest}\textbf{0.891} & \cellcolor{PaperBest}\textbf{0.000} & 46.4 \\
\bottomrule\end{tabular}
\end{table*}

\begin{table}[t]\centering
\caption{\tablelead{Response length and answer coverage across backbones.} Short-answer and explanation prompts use the same 904 queries from SQuAD-4, NQ-4, and Wikipedia birth year. Each backbone configuration uses one checkpoint for both instructions, which are represented in the multi-style recipe.}
\label{tab:prompt-cross-instruction}
\footnotesize
\setlength{\tabcolsep}{2.5pt}\renewcommand{\arraystretch}{1.10}
\setlength{\aboverulesep}{0pt}\setlength{\belowrulesep}{0pt}
\begin{tabular}{>{\raggedright\arraybackslash}p{\dimexpr 0.36000000\linewidth-2\tabcolsep\relax}>{\centering\arraybackslash}p{\dimexpr 0.16000000\linewidth-2\tabcolsep\relax}>{\centering\arraybackslash}p{\dimexpr 0.16000000\linewidth-2\tabcolsep\relax}>{\centering\arraybackslash}p{\dimexpr 0.16000000\linewidth-2\tabcolsep\relax}>{\centering\arraybackslash}p{\dimexpr 0.16000000\linewidth-2\tabcolsep\relax}}\toprule
\rowcolor{PaperHeader}
& \multicolumn{2}{c}{Short answer} & \multicolumn{2}{c}{Explanation} \\
\rowcolor{PaperHeader}
Method & Mean words & R-L & Mean words & R-L \\\midrule
\rowcolor{PaperGroup}\multicolumn{5}{l}{\textit{Qwen3-4B}} \\
ICL & 7.9 & 0.915 & 48.7 & 0.952 \\
D2L & 14.0 & 0.454 & 26.0 & 0.485 \\
\methodname{} & 5.3 & 0.844 & 46.8 & 0.875 \\
\rowcolor{PaperGroup}\multicolumn{5}{l}{\textit{Gemma-2-2B}} \\
ICL & 2.7 & 0.869 & 28.2 & 0.918 \\
D2L & 2.5 & 0.405 & 7.2 & 0.454 \\
\methodname{} & 2.2 & 0.816 & 27.0 & 0.859 \\
\bottomrule\end{tabular}
\end{table}

Table~\ref{tab:prompt-seen} compares response length and answer quality across instructions. Table~\ref{tab:prompt-cross-instruction} extends the comparison to Gemma. The direction depends on the writer: Gemma D2L stays much briefer than its ICL reference under explanation, whereas Qwen D2L is verbose under a short-answer request. \ours{} on Gemma expands from 2.2 to 27.0 words, with recall increasing from 0.816 to 0.859. ICL reaches 28.2 words and 0.918 recall, versus 7.2 words and 0.454 for Gemma D2L. \ours{}'s short-response fraction falls from 0.94 to approximately zero under explanation. This replicates response adaptation with the complete method.

\paragraph{Normalized response-length span.}
We define response-length span as the maximum minus minimum task-macro mean word count across the tested instructions, normalized by the corresponding ICL span. Qwen uses the four instructions and 13 tasks in Table~\ref{tab:prompt-seen}, yielding 31.2\% for D2L and 89.9\% for \ours{}. Gemma uses the two instructions and three tasks in Table~\ref{tab:prompt-cross-instruction}, yielding 18.4\% and 97.5\%, respectively.

\begin{table*}[!tp]\centering
\caption{\tablelead{Format compliance and answer quality under held-out formats.} JSON and prefix use four matched tasks. Uppercase uses reference-eligible samples from three tasks. Content metrics score extracted answers independently of format compliance.}
\label{tab:prompt-heldout-detail}
\small
\setlength{\tabcolsep}{3pt}\renewcommand{\arraystretch}{1.10}
\setlength{\aboverulesep}{0pt}\setlength{\belowrulesep}{0pt}
\begin{tabular}{>{\raggedright\arraybackslash}p{\dimexpr 0.28000000\textwidth-2\tabcolsep\relax}>{\centering\arraybackslash}p{\dimexpr 0.14400000\textwidth-2\tabcolsep\relax}>{\centering\arraybackslash}p{\dimexpr 0.14400000\textwidth-2\tabcolsep\relax}>{\centering\arraybackslash}p{\dimexpr 0.14400000\textwidth-2\tabcolsep\relax}>{\centering\arraybackslash}p{\dimexpr 0.14400000\textwidth-2\tabcolsep\relax}>{\centering\arraybackslash}p{\dimexpr 0.14400000\textwidth-2\tabcolsep\relax}}\toprule
\rowcolor{PaperHeader}
Method & \shortstack{Format\\success} & F1 & R-L & EM & Joint \\\midrule
\rowcolor{PaperGroup}\multicolumn{6}{l}{\textit{Uppercase answer}} \\
ICL & 0.954 & 0.450 & 0.863 & 0.190 & 0.186 \\
D2L & 0.907 & 0.137 & 0.385 & 0.009 & 0.009 \\
\methodname{} & \cellcolor{PaperBest}\textbf{0.915} & \cellcolor{PaperBest}\textbf{0.442} & \cellcolor{PaperBest}\textbf{0.636} & \cellcolor{PaperBest}\textbf{0.234} & \cellcolor{PaperBest}\textbf{0.208} \\
\addlinespace[3pt]
\rowcolor{PaperGroup}\multicolumn{6}{l}{\textit{JSON object}} \\
ICL & 0.993 & 0.724 & 0.905 & 0.603 & 0.603 \\
D2L & 0.845 & 0.291 & 0.447 & 0.173 & 0.173 \\
\methodname{} & \cellcolor{PaperBest}\textbf{0.998} & \cellcolor{PaperBest}\textbf{0.710} & \cellcolor{PaperBest}\textbf{0.826} & \cellcolor{PaperBest}\textbf{0.569} & \cellcolor{PaperBest}\textbf{0.569} \\
\addlinespace[3pt]
\rowcolor{PaperGroup}\multicolumn{6}{l}{\textit{ANSWER: prefix}} \\
ICL & 0.997 & 0.421 & 0.932 & 0.229 & 0.229 \\
D2L & 0.746 & 0.169 & 0.492 & 0.035 & 0.035 \\
\methodname{} & \cellcolor{PaperBest}\textbf{0.999} & \cellcolor{PaperBest}\textbf{0.487} & \cellcolor{PaperBest}\textbf{0.836} & \cellcolor{PaperBest}\textbf{0.295} & \cellcolor{PaperBest}\textbf{0.295} \\
\bottomrule\end{tabular}
\end{table*}

\paragraph{Response evaluations.}
The four-prompt sweep measures response-length adaptation alongside reference coverage. Under held-out formats, \ours{} improves joint format compliance and answer correctness over D2L across all three families. Relative to ICL, joint success is higher for uppercase and prefix instructions and lower for JSON, while reference coverage remains lower across the three formats. These comparisons distinguish compliance with an output format from the quality of the answer expressed within it.

\subsection{Ablation Studies}
Table~\ref{tab:current-components} summarizes writer and supervision ablations. Component definitions and matched training settings appear in Appendix~\ref{app:new-method}.

\paragraph{Teacher recipes across response instructions.}
\begin{table*}[!t]\centering
\caption{\tablelead{Teacher-recipe comparison across response instructions.} D2L prompt denotes single-style supervision. All configurations use \ours{} with the same source documents, queries, encoder architecture, and training-step budget. Panel (a) reports Qwen3-4B macro averages over the 17-task evaluation suite. Panel (b) follows the cohorts and scoring of Table~\ref{tab:prompt-heldout-detail}.}
\label{tab:teacher-format}
\small
\setlength{\tabcolsep}{3pt}\renewcommand{\arraystretch}{1.10}
\setlength{\aboverulesep}{0pt}\setlength{\belowrulesep}{0pt}
\begin{tabular}{>{\raggedright\arraybackslash}p{\dimexpr .17\textwidth-2\tabcolsep\relax}>{\raggedright\arraybackslash}p{\dimexpr .23\textwidth-2\tabcolsep\relax}*{4}{>{\centering\arraybackslash}p{\dimexpr .15\textwidth-2\tabcolsep\relax}}}\toprule
\rowcolor{PaperGroup}\multicolumn{6}{l}{\textit{(a) Teacher setting and answering instruction}} \\
\rowcolor{PaperHeader}
Instruction & Teacher setting & \shortstack{Mean\\words} & F1 & R-L & EM \\\midrule
Short answer & D2L prompt & 19.6 & 0.307 & \cellcolor{PaperBest}\textbf{0.777} & 0.088 \\
 & Multi-style & 7.2 & \cellcolor{PaperBest}\textbf{0.663} & 0.774 & \cellcolor{PaperBest}\textbf{0.510} \\
\addlinespace[3pt]
Entity & D2L prompt & 9.6 & 0.579 & 0.734 & 0.431 \\
 & Multi-style & 3.5 & \cellcolor{PaperBest}\textbf{0.706} & \cellcolor{PaperBest}\textbf{0.739} & \cellcolor{PaperBest}\textbf{0.572} \\
\addlinespace[3pt]
Sentence & D2L prompt & 15.5 & 0.254 & 0.771 & \cellcolor{PaperBest}\textbf{0.003} \\
 & Multi-style & 13.2 & \cellcolor{PaperBest}\textbf{0.273} & \cellcolor{PaperBest}\textbf{0.793} & \cellcolor{PaperBest}\textbf{0.003} \\
\addlinespace[3pt]
Query only & D2L prompt & 37.2 & 0.207 & 0.796 & 0.002 \\
 & Multi-style & 26.0 & \cellcolor{PaperBest}\textbf{0.246} & \cellcolor{PaperBest}\textbf{0.808} & \cellcolor{PaperBest}\textbf{0.006} \\
\addlinespace[3pt]
Explanation & D2L prompt & 47.7 & \cellcolor{PaperBest}\textbf{0.094} & 0.795 & 0.000 \\
 & Multi-style & 49.3 & 0.087 & \cellcolor{PaperBest}\textbf{0.819} & 0.000 \\
\bottomrule\end{tabular}
\par\medskip
\begin{tabular}{>{\raggedright\arraybackslash}p{\dimexpr .46\textwidth-2\tabcolsep\relax}*{3}{>{\centering\arraybackslash}p{\dimexpr .18\textwidth-2\tabcolsep\relax}}}\toprule
\rowcolor{PaperGroup}\multicolumn{4}{l}{\textit{(b) Joint success under held-out formats}} \\
\rowcolor{PaperHeader}
Teacher setting & Uppercase & JSON & Prefix \\\midrule
D2L prompt & 0.023 & 0.356 & 0.013 \\
D2L prompt, w/o $\mathcal{L}_{\mathrm{retain}}$ & 0.018 & 0.345 & 0.019 \\
Multi-style & \cellcolor{PaperBest}\textbf{0.208} & \cellcolor{PaperBest}\textbf{0.569} & \cellcolor{PaperBest}\textbf{0.295} \\
\bottomrule\end{tabular}
\end{table*}

Table~\ref{tab:current-components}b compares multi-style supervision with short-answer-only and D2L-prompt single-style controls under matched training data and budget. Multi-style supervision improves F1 and EM over both controls. R-L is higher than short-answer-only supervision and similar to D2L-prompt supervision. \looseness=-1
Table~\ref{tab:teacher-format}a compares D2L-prompt and multi-style supervision under five answering instructions on the same 17 tasks. The original-prompt model benefits from the more explicit entity instruction, while the multi-style recipe retains higher short-answer F1 and EM at similar reference coverage. Both models produce longer responses when explanations are requested. The multi-style recipe preserves this behavior and increases reference coverage under sentence, query-only, and explanation prompts. We report these answers' length and coverage, with F1 and EM scored against unchanged short references.

Table~\ref{tab:teacher-format}b compares the same checkpoints under held-out format instructions. The multi-style recipe improves joint success across all three families. Both teacher recipes and the no-retention control have similar format compliance.

\paragraph{Retention on Unrelated Inputs}
\label{app:retention-behavior}
We evaluate prediction drift on 40 unrelated biography and dialogue passages, using their first 128 tokens. Updates are constructed for 64 SQuAD-4 and 64 Wikipedia birth-year queries. KL measures divergence from the unmodified backbone's predictive distribution. The matched D2L-prompt checkpoints differ only in retention weight. Retention reduces KL from 0.2818 to 0.0265 and keeps perplexity close to the base model, while QA F1 changes from 0.313 to 0.307. The multi-style model also shows low drift. Retention thus limits interference without improving QA accuracy.

\paragraph{Unrelated mathematical reasoning.}
We evaluate the first 200 GSM8K questions using the same model and standard GSM8K in-context examples in both conditions. The reference uses no generated LoRA; the comparison loads LoRA generated from documents unrelated to GSM8K while keeping the answering prompt identical. Accuracy is 94\% without LoRA and 92\% with LoRA, a two-percentage-point decrease. Thus, generated document memory largely preserves performance on this unrelated reasoning sample. This comparison evaluates the complete model, complementing the retention-weight ablation above.

\par\medskip\noindent\begin{minipage}{\linewidth}\centering
\captionof{table}{\tablelead{Retention on unrelated text.} Base-model perplexity is 13.05. QA F1 is the 17-task macro average.}
\footnotesize\setlength{\tabcolsep}{3pt}
\begin{tabular}{lrrrr}\toprule
\rowcolor{PaperHeader}
Teacher & $\lambda$ & KL & PPL & QA F1 \\\midrule
D2L-prompt & 0 & 0.2818 & 15.08 & 0.313 \\
D2L-prompt & 0.3 & 0.0265 & 12.93 & 0.307 \\
\midrule
Multi-style & 0.3 & 0.0238 & 12.90 & 0.663 \\
\bottomrule\end{tabular}
\end{minipage}\par\medskip

\subsection{Hybrid Answering}
\label{app:kv-results}
Figure~\ref{fig:decoding-state-quality} compares KV-only answering with the addition of generated LoRA or Memory Tokens KV across retained-context budgets. At $B=48,128,512$, both added-memory forms use 6.75\,MiB and share the selected document cache. LoRA gives higher F1 and EM with both selectors at every tested shared budget, whereas Memory Tokens KV gives higher R-L. Both improve on KV-only answering. This tests complementarity beyond standalone accuracy (Table~\ref{tab:current-components}c).

At $B=48$, SnapKV and DropKV with \ours{} each use 13.50\,MiB of document-derived state. KV-only answering at $B=128$ uses 18.00\,MiB, yet has lower F1 and EM with comparable recall. These sizes follow Figure~\ref{fig:decoding-state-quality} and exclude prompt states.

\paragraph{Hybrid Controls and Evidence Interventions}
\begin{table*}[t]\centering
\caption{\tablelead{\ours{} gains under the same answering pipeline.} Macro averages over 17 matched tasks. Both KV methods use $B=48$. Within each KV method, all conditions use the same selected document KV and process the instruction and query over that cache. The lower block uses documents with multiple queries.}
\label{tab:hybrid-controls}
\footnotesize
\setlength{\tabcolsep}{3pt}\renewcommand{\arraystretch}{1.10}
\begin{tabular}{@{}lrrrrrr@{}}\toprule
\rowcolor{PaperHeader}
Configuration & Snap F1 & R-L & EM & Drop F1 & R-L & EM \\\midrule
KV only & 0.636 & 0.789 & 0.485 & 0.620 & 0.780 & 0.472 \\
KV + \ours{} & 0.696 & 0.816 & 0.545 & 0.705 & 0.825 & 0.555 \\
KV + \ours{} (other document) & 0.573 & 0.720 & 0.424 & 0.570 & 0.717 & 0.426 \\
\ours{} only & 0.663 & 0.774 & 0.510 & 0.663 & 0.774 & 0.510 \\
\midrule
\rowcolor{PaperGroup}\multicolumn{7}{l}{\textit{Same-document query swap: matched subset of eight tasks}} \\
KV only & 0.764 & 0.908 & 0.605 & 0.741 & 0.906 & 0.588 \\
KV + \ours{} (current query) & 0.815 & 0.930 & 0.670 & 0.820 & 0.937 & 0.677 \\
KV + \ours{} (other query) & 0.665 & 0.794 & 0.512 & 0.671 & 0.801 & 0.532 \\
\bottomrule\end{tabular}
\end{table*}

\paragraph{Matched answering pipeline.}
With the answering pipeline fixed, adding \ours{} raises F1 by 0.060 for SnapKV and 0.085 for DropKV. An update from another document or query scores below KV-only answering (Table~\ref{tab:hybrid-controls}).

\begin{table*}[t]\centering
\caption{\tablelead{Answer-position retention control.} Seven tasks with answer-matching KV positions prioritized within budget. Each pair shares the retained cache and query replay.}
\label{tab:hybrid-retention}
\footnotesize
\setlength{\tabcolsep}{3pt}\renewcommand{\arraystretch}{1.10}
\begin{tabular}{@{}lrrrrrrr@{}}\toprule
\rowcolor{PaperHeader}
 & & \multicolumn{3}{c}{KV only} & \multicolumn{3}{c}{KV + \ours{}} \\
\rowcolor{PaperHeader}
Method & $B$ & F1 & R-L & EM & F1 & R-L & EM \\\midrule
SnapKV & 16 & 0.669 & 0.847 & 0.556 & 0.749 & 0.861 & 0.630 \\
SnapKV & 48 & 0.779 & 0.925 & 0.636 & 0.811 & 0.930 & 0.673 \\
DropKV & 16 & 0.550 & 0.756 & 0.452 & 0.700 & 0.815 & 0.597 \\
DropKV & 48 & 0.759 & 0.911 & 0.615 & 0.810 & 0.924 & 0.672 \\
\bottomrule\end{tabular}
\end{table*}

\paragraph{Answer-position retention.}
Table~\ref{tab:hybrid-retention} reports the seven-task macro averages. Restricting the analysis to the 2,068 queries with retained answer-matching positions preserves the effect. At $B=48$, adapter use raises F1 from 0.781 to 0.812 for SnapKV and from 0.760 to 0.809 for DropKV.

\begin{table*}[t]\centering
\caption{\tablelead{Counterfactual evidence conflicts.} Cells report the fractions of responses containing the original ($X$) and counterfactual ($Y$) answers. KV and \ours{} use independently chosen documents for each query.}
\label{tab:hybrid-conflict}
\footnotesize
\setlength{\tabcolsep}{3pt}\renewcommand{\arraystretch}{1.10}
\begin{tabular}{@{}lrrrrr@{}}\toprule
\rowcolor{PaperHeader}
Method & $B$ & KV source & KV only & + \ours{} ($X$) & + \ours{} ($Y$) \\\midrule
\rowcolor{PaperGroup}\multicolumn{6}{l}{\textit{FactSet-Homo}} \\
SnapKV & 16 & X & 0.54 / 0.00 & 0.87 / 0.00 & 0.45 / 0.06 \\
SnapKV & 16 & Y & 0.00 / 0.54 & 0.03 / 0.41 & 0.00 / 0.87 \\
SnapKV & 48 & X & 0.95 / 0.00 & 0.99 / 0.00 & 0.86 / 0.02 \\
SnapKV & 48 & Y & 0.00 / 0.93 & 0.01 / 0.87 & 0.00 / 0.97 \\
DropKV & 16 & X & 0.93 / 0.00 & 0.99 / 0.00 & 0.82 / 0.05 \\
DropKV & 16 & Y & 0.00 / 0.92 & 0.09 / 0.81 & 0.00 / 0.99 \\
DropKV & 48 & X & 1.00 / 0.00 & 1.00 / 0.00 & 0.92 / 0.02 \\
DropKV & 48 & Y & 0.00 / 1.00 & 0.01 / 0.93 & 0.00 / 1.00 \\
\rowcolor{PaperGroup}\multicolumn{6}{l}{\textit{Wikipedia year}} \\
SnapKV & 16 & X & 0.33 / 0.00 & 0.91 / 0.00 & 0.05 / 0.66 \\
SnapKV & 16 & Y & 0.07 / 0.30 & 0.76 / 0.03 & 0.00 / 0.87 \\
SnapKV & 48 & X & 0.98 / 0.00 & 0.98 / 0.00 & 0.71 / 0.13 \\
SnapKV & 48 & Y & 0.00 / 0.97 & 0.17 / 0.67 & 0.00 / 0.97 \\
DropKV & 16 & X & 0.56 / 0.01 & 0.93 / 0.00 & 0.12 / 0.57 \\
DropKV & 16 & Y & 0.07 / 0.49 & 0.63 / 0.12 & 0.00 / 0.95 \\
DropKV & 48 & X & 0.99 / 0.00 & 0.99 / 0.00 & 0.85 / 0.08 \\
DropKV & 48 & Y & 0.00 / 0.98 & 0.05 / 0.85 & 0.00 / 0.97 \\
\bottomrule\end{tabular}
\end{table*}

\paragraph{Counterfactual conflicts.}
Table~\ref{tab:hybrid-conflict} reports original-answer and counterfactual-answer containment separately. At small budgets on Wikipedia, changing only the adapter source substantially shifts which fact appears in the response. At larger budgets, the cached source remains dominant, while a conflicting adapter can reduce its answer containment. These interventions show both factual supplementation and interference between the two memory forms, with their relative influence varying with the KV budget.

\subsection{Efficiency}
\label{app:efficiency-measured}
\begin{table*}[!tp]\centering
\caption{\tablelead{Preparation and decoding efficiency.} Qwen3-4B on one H200, grouped by output length. Document prefill is paid once. Per-query preparation ends at the first answer token. State excludes model weights and document caches retained for future queries. Throughput uses batch 16; OOM denotes out of memory.}
\label{tab:efficiency}
\footnotesize
\setlength{\tabcolsep}{3pt}\renewcommand{\arraystretch}{1.10}
\setlength{\aboverulesep}{0pt}\setlength{\belowrulesep}{0pt}
\begin{tabular}{>{\raggedright\arraybackslash}p{\dimexpr 0.065\textwidth-2\tabcolsep\relax}>{\raggedright\arraybackslash}p{\dimexpr 0.225\textwidth-2\tabcolsep\relax}>{\centering\arraybackslash}p{\dimexpr 0.105\textwidth-2\tabcolsep\relax}>{\centering\arraybackslash}p{\dimexpr 0.105\textwidth-2\tabcolsep\relax}>{\centering\arraybackslash}p{\dimexpr 0.115\textwidth-2\tabcolsep\relax}>{\centering\arraybackslash}p{\dimexpr 0.115\textwidth-2\tabcolsep\relax}>{\centering\arraybackslash}p{\dimexpr 0.135\textwidth-2\tabcolsep\relax}>{\centering\arraybackslash}p{\dimexpr 0.135\textwidth-2\tabcolsep\relax}}\toprule
\rowcolor{PaperGroup}\multicolumn{8}{l}{\textit{(a) 64 output tokens: document lengths and KV budgets}} \\
\rowcolor{PaperHeader}
& & \multicolumn{2}{c}{State (MiB)} & Prefill & Prepare & Decode & Throughput \\
\rowcolor{PaperHeader}
Doc. & Method & Total & Document & (ms) & (ms) & \shortstack{batch 1\\(ms/tok)} & \shortstack{batch 16\\(tok/s)} \\\midrule
\multirow{6}{*}{4K} & ICL (full document) & 581.9 & 576.4 & 279 & 29.6 & 22.38 & 512 \\
 & SnapKV, $B{=}48$ & 12.7 & 6.8 & 279 & 47.0 & 22.35 & 627 \\
 & DropKV, $B{=}48$ & 12.7 & 6.8 & 279 & 59.7 & 23.02 & 644 \\
 & SnapKV, $B{=}128$ & 23.9 & 18.0 & 279 & 47.4 & 22.42 & 649 \\
 & DropKV, $B{=}128$ & 23.9 & 18.0 & 279 & 60.4 & 22.73 & 595 \\
 & \methodname{} & 12.5 & 6.8 & 178 & 123.0 & 22.33 & 598 \\
\midrule
\multirow{6}{*}{16K} & ICL (full document) & 2309.9 & 2304.4 & 592 & 35.4 & 22.49 & 178 \\
 & SnapKV, $B{=}48$ & 12.7 & 6.8 & 592 & 71.6 & 22.32 & 598 \\
 & DropKV, $B{=}48$ & 12.7 & 6.8 & 592 & 114.5 & 22.50 & 597 \\
 & SnapKV, $B{=}128$ & 23.9 & 18.0 & 592 & 71.8 & 22.52 & 645 \\
 & DropKV, $B{=}128$ & 23.9 & 18.0 & 592 & 114.4 & 22.88 & 637 \\
 & \methodname{} & 12.5 & 6.8 & 669 & 123.2 & 22.07 & 596 \\
\midrule
\multirow{6}{*}{32K} & ICL (full document) & 4613.9 & 4608.4 & 1630 & 45.2 & 22.92 & 95 \\
 & SnapKV, $B{=}48$ & 12.7 & 6.8 & 1630 & 102.0 & 22.75 & 651 \\
 & DropKV, $B{=}48$ & 12.7 & 6.8 & 1630 & 185.6 & 22.67 & 635 \\
 & SnapKV, $B{=}128$ & 23.9 & 18.0 & 1630 & 101.7 & 22.77 & 644 \\
 & DropKV, $B{=}128$ & 23.9 & 18.0 & 1630 & 184.1 & 22.72 & 605 \\
 & \methodname{} & 12.5 & 6.8 & 1796 & 171.3 & 22.59 & 526 \\
\midrule
\multirow{6}{*}{64K} & ICL (full document) & 9221.9 & 9216.4 & 5185 & 61.7 & 29.70 & OOM \\
 & SnapKV, $B{=}48$ & 12.7 & 6.8 & 5185 & 167.6 & 22.98 & 633 \\
 & DropKV, $B{=}48$ & 12.7 & 6.8 & 5185 & 334.5 & 22.84 & 637 \\
 & SnapKV, $B{=}128$ & 23.9 & 18.0 & 5185 & 165.4 & 22.87 & 630 \\
 & DropKV, $B{=}128$ & 23.9 & 18.0 & 5185 & 330.9 & 23.77 & 604 \\
 & \methodname{} & 12.5 & 6.8 & 5535 & 268.7 & 22.98 & 564 \\
\midrule
\multirow{6}{*}{128K} & ICL (full document) & 18437.9 & 18432.4 & 18038 & 97.3 & 49.37 & OOM \\
 & SnapKV, $B{=}48$ & 12.7 & 6.8 & 18038 & 295.1 & 23.07 & 607 \\
 & DropKV, $B{=}48$ & 12.7 & 6.8 & 18038 & 619.7 & 23.08 & 614 \\
 & SnapKV, $B{=}128$ & 23.9 & 18.0 & 18038 & 294.3 & 23.57 & 594 \\
 & DropKV, $B{=}128$ & 23.9 & 18.0 & 18038 & 621.0 & 23.61 & 597 \\
 & \methodname{} & 12.5 & 6.8 & 19724 & 476.6 & 22.96 & 583 \\
\bottomrule\end{tabular}
\end{table*}

\begin{table*}[!tp]\ContinuedFloat\centering
\caption[]{\tablelead{Preparation and decoding efficiency (continued).} KV baselines in (b) retain up to 48 document positions. State in (b) is measured at the first answer token. Doc. peak $\Delta$ is the prefill peak allocation increment above the phase-start baseline; -- denotes no separate document phase. In (c), query latency includes preparation and batch-one generation after document caching, and state includes the generated output.}
\label{tab:efficiency-cached-full}\label{tab:efficiency-output-length}
\footnotesize
\setlength{\tabcolsep}{3pt}\renewcommand{\arraystretch}{1.10}
\setlength{\aboverulesep}{0pt}\setlength{\belowrulesep}{0pt}
\begin{tabular}{>{\raggedright\arraybackslash}p{\dimexpr 0.065\textwidth-2\tabcolsep\relax}>{\raggedright\arraybackslash}p{\dimexpr 0.25\textwidth-2\tabcolsep\relax}>{\centering\arraybackslash}p{\dimexpr 0.12\textwidth-2\tabcolsep\relax}>{\centering\arraybackslash}p{\dimexpr 0.12\textwidth-2\tabcolsep\relax}>{\centering\arraybackslash}p{\dimexpr 0.135\textwidth-2\tabcolsep\relax}>{\centering\arraybackslash}p{\dimexpr 0.14\textwidth-2\tabcolsep\relax}>{\centering\arraybackslash}p{\dimexpr 0.17\textwidth-2\tabcolsep\relax}}\toprule
\rowcolor{PaperGroup}\multicolumn{7}{l}{\textit{(b) 256 output tokens: document caching and per-query costs}} \\
\rowcolor{PaperHeader}
 & & \multicolumn{2}{c}{Preparation (ms)} & Req. state & Throughput & Doc. peak $\Delta$ \\
\rowcolor{PaperHeader}
Doc. & Method & Document & Query & (MiB) & \shortstack{batch 16\\(tok/s)} & (MiB) \\\midrule
\multirow{7}{*}{1K} & ICL (full document) & 36.3 & 29.1 & 149.91 & 630 & 222.2 \\
 & SnapKV & 36.3 & 46.3 & 12.66 & 624 & 222.2 \\
 & DropKV & 36.3 & 51.9 & 12.66 & 626 & 222.2 \\
 & CCM & 171.9 & 29.5 & 12.52 & 607 & 475.2 \\
 & 500xCompressor & 178.0 & 30.5 & 12.52 & 628 & 350.0 \\
 & Re-encoding control & 0.0 & 200.3 & 12.52 & 626 & -- \\
 & \methodname{} & 100.0 & 135.8 & 12.52 & 575 & 251.6 \\
\midrule
\multirow{7}{*}{4K} & ICL (full document) & 114.6 & 29.4 & 581.91 & 512 & 886.7 \\
 & SnapKV & 114.6 & 47.5 & 12.66 & 605 & 886.7 \\
 & DropKV & 114.6 & 60.2 & 12.66 & 627 & 886.7 \\
 & CCM & 243.6 & 28.4 & 12.52 & 634 & 1354.5 \\
 & 500xCompressor & 246.2 & 29.0 & 12.52 & 648 & 1231.1 \\
 & Re-encoding control & 0.0 & 277.2 & 12.52 & 651 & -- \\
 & \methodname{} & 157.6 & 131.4 & 12.52 & 596 & 887.8 \\
\midrule
\multirow{7}{*}{16K} & ICL (full document) & 585.8 & 35.9 & 2309.91 & 178 & 3544.8 \\
 & SnapKV & 585.8 & 72.1 & 12.66 & 635 & 3544.8 \\
 & DropKV & 585.8 & 114.4 & 12.66 & 616 & 3544.8 \\
 & CCM & 758.7 & 29.3 & 12.52 & 632 & 5006.5 \\
 & 500xCompressor & 824.0 & 29.0 & 12.52 & 634 & 4883.1 \\
 & Re-encoding control & 0.0 & 850.3 & 12.52 & 628 & -- \\
 & \methodname{} & 666.3 & 128.1 & 12.52 & 583 & 3546.1 \\
\midrule
\multirow{7}{*}{32K} & ICL (full document) & 1648.9 & 45.4 & 4613.91 & 95 & 7088.9 \\
 & SnapKV & 1648.9 & 102.6 & 12.66 & 606 & 7088.9 \\
 & DropKV & 1648.9 & 185.6 & 12.66 & 601 & 7088.9 \\
 & CCM & 1958.8 & 29.4 & 12.52 & 618 & 9875.8 \\
 & 500xCompressor & 2105.5 & 29.5 & 12.52 & 617 & 9752.5 \\
 & Re-encoding control & 0.0 & 2130.4 & 12.52 & 612 & -- \\
 & \methodname{} & 1820.6 & 174.6 & 12.52 & 560 & 7090.5 \\
\bottomrule\end{tabular}
\par\medskip
\begin{tabular}{>{\raggedright\arraybackslash}p{\dimexpr 0.065\textwidth-2\tabcolsep\relax}>{\raggedright\arraybackslash}p{\dimexpr 0.275\textwidth-2\tabcolsep\relax}>{\centering\arraybackslash}p{\dimexpr 0.140\textwidth-2\tabcolsep\relax}>{\centering\arraybackslash}p{\dimexpr 0.180\textwidth-2\tabcolsep\relax}>{\centering\arraybackslash}p{\dimexpr 0.180\textwidth-2\tabcolsep\relax}>{\centering\arraybackslash}p{\dimexpr 0.160\textwidth-2\tabcolsep\relax}}\toprule
\rowcolor{PaperGroup}\multicolumn{6}{l}{\textit{(c) Output-length sweep: query latency and final request state}} \\
\rowcolor{PaperHeader}
 & & Output & Query latency & Throughput & State \\
\rowcolor{PaperHeader}
Doc. & Method & (tokens) & \shortstack{batch 1\\(ms)} & \shortstack{batch 16\\(tok/s)} & (MiB) \\\midrule
\multirow{9}{*}{4K} & \multirow{3}{*}{ICL (full document)} & 16 & 369.4 & 519 & 584.02 \\
 &  & 256 & 5789.2 & 512 & 617.77 \\
 &  & 1024 & 23214.2 & 484 & 725.77 \\
\cmidrule{2-6}
 & \multirow{3}{*}{Re-encoding control} & 16 & 619.7 & 654 & 14.62 \\
 &  & 256 & 6220.4 & 651 & 48.38 \\
 &  & 1024 & 23891.6 & 651 & 156.38 \\
\cmidrule{2-6}
 & \multirow{3}{*}{\methodname{}} & 16 & 468.4 & 600 & 14.62 \\
 &  & 256 & 6039.4 & 596 & 48.38 \\
 &  & 1024 & 24272.3 & 573 & 156.38 \\
\midrule
\multirow{9}{*}{32K} & \multirow{3}{*}{ICL (full document)} & 16 & 400.1 & 96 & 4616.02 \\
 &  & 256 & 6167.0 & 95 & 4649.77 \\
 &  & 1024 & 24330.5 & 94 & 4757.77 \\
\cmidrule{2-6}
 & \multirow{3}{*}{Re-encoding control} & 16 & 2513.1 & 620 & 14.62 \\
 &  & 256 & 8261.3 & 612 & 48.38 \\
 &  & 1024 & 26505.7 & 611 & 156.38 \\
\cmidrule{2-6}
 & \multirow{3}{*}{\methodname{}} & 16 & 544.2 & 557 & 14.62 \\
 &  & 256 & 6335.9 & 560 & 48.38 \\
 &  & 1024 & 24990.5 & 561 & 156.38 \\
\bottomrule\end{tabular}
\end{table*}

\looseness=-1
Table~\ref{tab:efficiency} groups measurements by output length. Panel (a) reports preparation, state, and decoding costs with 64 output tokens. Panel (b) compares document caching and per-query costs with 256 output tokens. Panel (c) varies output length at fixed document lengths and includes answer generation. Document caching is paid once per document, while each query pays memory construction and preparation through the first answer token.

\paragraph{Document-prefill memory.}
Table~\ref{tab:efficiency-cached-full}b reports the median document-prefill peak allocation increment above GPU memory allocated immediately before that phase, measured with PyTorch peak-allocation counters after warmup. It includes allocations made during document prefill, but excludes memory already resident at the start of that phase. It is not the total deployment peak, nor the peak for per-query LoRA construction.

\end{document}